\documentclass[conference]{IEEEtran}
\usepackage{times}
\usepackage{helvet}
\usepackage{courier}

\usepackage{color}
\usepackage{url}
\usepackage{amsfonts}
\usepackage{amssymb}
\usepackage{amsmath}
\usepackage{stmaryrd}
\usepackage{amsthm}
\usepackage{algorithm}
\usepackage{algpseudocode}
\usepackage{comment}
\usepackage{graphicx}
\usepackage{caption}
\usepackage{subcaption}
\usepackage{float}
\usepackage{multirow}
\usepackage{tabularx}
\usepackage{subcaption}
\usepackage{cite}

\newcommand{\Mv}[1]{\mathbf{#1}}
\newcommand{\Mt}[1]{\mathtt{#1}}
\newcommand{\Mc}[1]{\mathcal{#1}}
\newcommand{\Mvh}[1]{\hat{\mathbf{#1}}}

\newcommand{\argmin}[1]{\underset{#1}{\arg\min}\quad}

\newcommand{\argmax}[1]{\underset{#1}{\arg\max}\;}
\newcommand{\td}[1]{\tilde{#1}}
\newcommand{\setre}[1]{\renewcommand{\algorithmicrequire}{\textbf{#1}}}
\renewcommand{\algorithmicrequire}{\textbf{Input:}}

\newcommand{\Mvt}[1]{\mathbf{\td{#1}}}

\begin{document}
%
\title{Can Active Learning Experience Be Transferred?}
\author{
\IEEEauthorblockN{Hong-Min Chu}
            \IEEEauthorblockA{Department of Computer Science and\\
                    Information Engineering, \\ 
                    National Taiwan University \\
            E-mail: r04922031@csie.ntu.edu.tw}
\and
\IEEEauthorblockN{Hsuan-Tien Lin}
            \IEEEauthorblockA{Department of Computer Science and \\
                    Information Engineering, \\ 
                    National Taiwan University \\
            E-mail: htlin@csie.ntu.edu.tw}
}
\maketitle
\begin{abstract}
\begin{quote}
Active learning is an important machine learning problem in reducing the human labeling effort. Current active learning strategies are designed from human knowledge, and are applied on each dataset in an immutable manner. In other words, experience about the usefulness of strategies cannot be updated and transferred to improve active learning on other datasets. This paper initiates a pioneering study on whether active learning experience can be transferred. We first propose a novel active learning model that linearly aggregates existing strategies. The linear weights can then be used to represent the active learning experience. We equip the model with the popular linear upper-confidence-bound (LinUCB) algorithm for contextual bandit to update the weights. Finally, we extend our model to transfer the experience across datasets with the technique of biased regularization. Empirical studies demonstrate that the learned experience not only is competitive with existing strategies on most single datasets, but also can be transferred across datasets to improve the performance on future learning tasks. 
\end{quote}
\end{abstract}

\section{Introduction} \label{intro}


In many machine learning applications, high-quality labels are costly to obtain \cite{active_cancer, active_information}. 
Active learning is a machine learning scenario that tries to reduce the labeling cost while still maintaining the performance of learned models by asking key labeling questions \cite{Survey}. 
Most current active learning algorithms are based on human knowledge about how to ask questions, and the knowledge is applied immutably on every dataset when conducting active learning.
A recent work \cite{ALBL} argued that any single active learning algorithm based on immutable human knowledge is unlikely to perform well on all datasets, and hence proposed to adaptively learn a probabilistic blending of a set of human-designed active learning algorithms.
The blending is learned within a single dataset via connecting  with multi-armed bandit learning.
Given the possibility to learn a decent blending of different pieces of human knowledge within a single dataset, our key thought is:
can the learned experience be transferred to other datasets to improve the performance of active learning? 

Our thought is related to how human beings learn to ask questions in real life. We do not just learn to ask questions within a single learning task; we instead accumulate experience in question-asking in past and current learning tasks and transfer the experience to future learning tasks. 
There are setups in machine learning that study how experience can be transferred to future tasks. 
The simplest setup is transfer learning~\cite{survey_transfer}, or inductive transfer. 
Transfer learning is about accumulating experience from one or several source tasks and applying the experience to a related target task.
Several attempts have been made in previous studies to improve the performance of active learning with transfer learning~\cite{AcTrak,transfer_active_stream,HATL}. 
However, all the algorithms proposed in these studies aim to transfer the experience of supervised or semi-supervised learning from the source tasks to the target task, 
and do not transfer the experience of active learning (question-asking).
Furthermore, the algorithms assume a shared feature space between different tasks, 
while experience transfer between heterogeneous active learning tasks is yet to be studied. 

Other related setups include never-ending learning and life-long learning. 
Never-ending learning is a rather general setup that defines how machines can learn like humans to transfer experience to different tasks in a self-supervised manner, 
and has been realized in a system for accumulating beliefs by reading continuously from the web~\cite{never_ending_learning}. 
Life-long learning~\cite{life_long_sentiment,life_long_system}, on the other hand, considers feeding the machines with a sequence of tasks with the hope of improving the performance on the next task in the sequence. 
The setup is similar to our thought but has been realized on only sentiment classification tasks~\cite{life_long_sentiment}. 

To the best of our knowledge, neither never-ending nor life-long learning has been carried out on active learning tasks. In fact, allowing the machine to mimic humans in life-long active learning is highly non-trivial, as experience that can be accumulated and transferred between heterogeneous active learning tasks is not well-defined, not to mention applying past experience to future learning tasks. 

In this paper, after introducing the cross-dataset (cross-task) active learning problem in Section~\ref{background}, 
we first propose a notion of machine experience that can be transferred across active learning tasks in Section~\ref{LSA}. 
The notion is based on encoding human knowledge of active learning via scoring functions of existing active learning algorithms, and representing machine experience as linear weights that combine the human knowledge.
Under the notion, existing active learning algorithms can be simply viewed as taking some special and immutable weights to combine the knowledge. 

Then, we improve existing active learning algorithms by designing a novel approach that adaptively update the linear weights during the active learning process. 
Inspired by the aforementioned work \cite{ALBL}, we connect our problem of updating the linear weights with contextual bandit learning. 
Based on the connection, we apply a state-of-the-art contextual bandit algorithm, \textit{Linear Upper-Confidence-Bound} (LinUCB) \cite{LinUCB}, to update the weights. 
The resulting approach effectively blends existing active learning algorithms towards better performance. 

We extend the proposed approach to allow the learned experience (weights) to be transferred across datasets in Section~\ref{Transfer_LSA}. The transferring extension is based on the idea of biased regularization that restricts the adaptive weights to be close to the past experience. The simple formulation of biased regularization can be seamlessly coupled with the LinUCB algorithm to form the transferring extension.

Empirical studies in Section~\ref{experiment} demonstrate that our approach is competitive to existing active learning algorithms. 
The results also indicate that the transferring extension effectively improves the learning performance of our approach with the experience learned
from both heterogeneous and homogeneous tasks,
thus demonstrating the usefulness of the learned experience. 
Finally, we conclude the possibility of transferring active learning
experience in Section~\ref{conclusion}.



\section{Background} \label{background}
In this work, we focus on a popular active learning setup called pool-based active learning \cite{Survey} for binary classification. 
Under the setup, an active learning algorithm is presented with a labeled pool and an unlabeled pool initially. 
We denote the labeled pool as $\mathcal{D}_l = \{(\Mv x_1,y_1), (\Mv x_2,y_2)...(\Mv x_{N_l},y_{N_l})\}$ and the unlabeled pool 
as $\mathcal{D}_u = \{\td{\Mv x}_1, \td{\Mv x}_2, ..., \td{\Mv x }_{N_u} \}$, 
where $\Mv x_i, \td{\Mv x}_{j} \in \mathbb{R}^d$, and $y_i \in \{+1,-1\}$. 
In general, the algorithm can only access a small~$\mathcal{D}_l$ in the beginning, while the size of $\mathcal{D}_u$ is relatively large.

With the initial $\mathcal{D}_l$, the algorithm calls some base model to learn a classifier $h_0$. Then, given a budget $T$, for each iteration $t=1,2,....,T$, the algorithm is allowed to query the label of an $\td{\Mv x}_{j} \in \mathcal{D}_u$ from some given labeling oracle. The instance-label pair $(\td{\Mv x}_{j}, y_j)$ will then be moved to $\mathcal{D}_l$, and the base model can be called with the enlarged $\mathcal{D}_l$ to learn a new classifier $h_t$. The goal of the algorithm is to make the performance of $h_1, h_2, ..., h_T$ as good as possible, where the performance will be measured with the test accuracy on a separate test set in this work.



We also study how active learning experience can be accumulated across datasets. In the setup of cross-dataset active learning,
we present the active learning algorithm with a sequence of datasets $(\mathcal{D}_l^{(1)}, \mathcal{D}_u^{(1)})$, $(\mathcal{D}_l^{(2)}, \mathcal{D}_u^{(2)})$, $\cdots$, 
$(\mathcal{D}_l^{(Q)}, \mathcal{D}_u^{(Q)})$, with the hope of improving the active learning performance along with the sequence like life-long learning \cite{life_long_sentiment,life_long_system}. 
More specifically, we hope that the experience accumulated from $(\mathcal{D}_l^{(1)}, \mathcal{D}_u^{(1)})$, $\cdots$, $(\mathcal{D}_l^{(q-1)}, \mathcal{D}_u^{(q-1)})$ 
can be exploited when conducting active learning on $(\mathcal{D}_l^{(q)}, \mathcal{D}_u^{(q)})$ for $q = 2, 3, \cdots, Q$.

Many active learning algorithms select $\td{\Mv x}_j$ from $ \mathcal{D}_u$ in iteration $t$ with a scoring function of instance $\td{\Mv x}$ subject to the current classifier $h_{t-1}$. For an algorithm $a$, we shall denote the scoring function as $s_{a}(\td{\Mv x}, h_{t-1})$, and assume that $a$ would query the label of $\td{\Mv x}_{j} = \argmax{\td{\Mv x} \in \mathcal{D}_u} s_{a}(\td{\Mv x}, h_{t-1})$. The scoring function measures the goodness of each instance, and reflects the strategy taken within the algorithm. 

A classic and intuitive strategy is called uncertainty sampling  \cite{uncertainty_start}, which queries the instance~$\td{\Mv x}_j$ 
that the classifier~$h_{t-1}$ is most uncertain with. \cite{uncertainty_svm} realizes the uncertainty sampling strategy
with a scoring function that computes the inverse distance from $\td{\Mv x}$ 
to the hyperplane of $h_{t-1}$ learned from Support Vector Machine (SVM).

Other works argue that uncertainty sampling only works well when $h_{t-1}$ is close enough to the ideal boundary, and may result in unsatisfactory performance 
when $h_{t-1}$ is not good enough \cite{hint_svm}. Representative sampling is a family of strategies, each based on a different scoring function, 
that tries to improve uncertainty sampling. For example, \cite{representative} applies $k$-means clustering and takes the inverse distance from $\td{\Mv x}$ to the cluster center
as the scoring function for representativeness, modulated by whether $\td{\Mv x}$ resides inside the margin of a SVM classifier~$h_{t-1}$.
\cite{DUAL} equips Gaussian distributions on top of $k$-means clustering to calculate representativeness, and proposes a scoring function that multiplies the uncertainty of $\td{\Mv x}$ by its representativeness. \cite{QUIRE} optimizes
a scoring function based on estimating the label assignments in a min-max view, and argues that the optimized scoring function covers both uncertainty and representativeness.

The strategies above embed our human knowledge of key labeling questions in the scoring functions. Several works~\cite{ALBL,COMB} also consider selecting the strategies adaptively for better performance, motivated by the fact that human-designed scoring functions cannot always match dataset characteristics and thus adaptive selection may be necessary. The state-of-the-art approach \textit{Active Learning By Learning}~\cite{ALBL} performs adaptive strategy selection by connecting the selection problem to bandit learning, and designs a learning-performance-based reward function to guide the bandit learner in selecting reasonable strategies probabilistically. The internal probability that each
strategy gets selected reflects the goodness of the strategy, and is updated
on the fly within the single dataset.

Recall that we aim to accumulate active learning experience across datasets. Human-designed scoring functions cannot help with so because they are generally immutable and cannot adaptively change with experience. A na{\"i}ve way of extending
current adaptive-selection approaches~\cite{ALBL,COMB}
for accumulating active learning experience
is to define the experience as the internal probability distribution for selections,
and then transfer the distribution to the next active learning task. Nevertheless,
as we shall see in Section~\ref{experiment}, 
the unstable nature of probabilistic choices makes the distribution too volatile to serve as robust active learning experience in practice. 



\section{Proposed Approach} \label{LSA}

In this section, we shall first introduce our notion of active learning experience. Then we propose a novel active learning approach, \textit{Linear Strategy Aggregation}, that queries an unlabeled instance and updates the experience simultaneously in each iteration.

\subsection{Notion of active learning experience}\label{sec:exp}

As introduced in Section~\ref{background}, the scoring functions of human-designed active learning algorithms represent pieces
of human knowledge about key labeling questions.
A proper way to combine different pieces of human knowledge, or namely different scoring functions, 
can then be naturally viewed as experience of active learning. 

More specifically, we consider combining, or \textit{blending}, the human-designed scoring functions to 
a new scoring  function for better performance, 
and define the blending parameters as experience. 
Note that current adaptive-selection approaches~\cite{ALBL} cannot fully match this novel definition, 
as they blend (via probabilistic selection) the recommended queries of the scoring functions instead of blending the scoring functions directly.

To take an initiative on the definition, we consider the simplest model where the scoring functions are blended linearly, and leave the possibility of using more sophisticated models as future directions. In particular,
given a set of scoring functions $\{s_1, s_2,\hdots, s_M\}$ from different human-designed strategies, we set the aggregated scoring function to be $\hat{s}(\td{\Mv x}, h_{t-1}) = \sum_{m=1}^M w_m s_m(\td{\Mv x}, h_{t-1})$. The weight vector $\mathbf{w} = (w_1, w_2, \cdots, w_M)$ then contains the blending parameters and 
serves as the experience that will be transferred.

\subsection{Linear Strategy Aggregation}\label{sec:LSA}

With the notion of experience established, we now introduce our proposed approach, \textit{Linear Strategy Aggregation}~(LSA).
LSA solves the task of adaptively updating the experience and querying the unlabeled instance $\Mvt x_j$ 
to maximize the active learning performance. 
Motivated by previous adaptive selection approaches~\cite{ALBL,COMB}, we design LSA via the connection between the task and   
a well-known adaptive learning problem of contextual bandit~\cite{bandit_start}. 
We will first discuss more details about the contextual bandit problem.

The setup of the contextual bandit problem is as follows~\cite{bandit_start}: 
a player is presented with $K$ actions and a budget $T$. 
In each iteration $t=1,\cdots,T$, the context vector~$\Mv z_{k,t}$ for each action $k \in \{1, 2, \cdots, K\}$ is provided, and a player is required to perform an action $k_t \in  \{1, 2, \cdots, K\}$. 
Once the action is performed, the corresponding reward $r_{k_t,t}$ is then revealed. The objective of the player is to maximize the cumulative reward. 
To maximize the cumulative reward, the player is typically required to 
balance between exploration (choosing actions that improve the estimation of reward) and exploitation (choosing actions with the highest estimated reward). 

Many algorithms for the contextual bandit problem have been studied in the literature~\cite{EXP4.P,LinRel,thompson,LinUCB}, 
and a family of them estimates the reward of an action through a linear model of the corresponding context~\cite{LinRel,thompson,LinUCB}.
A state-of-the-art algorithm of the family is called \textit{Linear Upper-Confidence-Bound} (LinUCB)~\cite{LinUCB}, 
which not only carries strong theoretical guarantees but also performs well on real-world tasks~\cite{LinUCB_news}.
Next, we take a closer look at LinUCB, and then apply it for LSA by connecting the contextual bandit problem back to active learning.

LinUCB maintains the weight vector $\Mv w_t$ of the linear model
to be the ridge regression solution from the context vectors to the observed rewards.
Specifically, before each iteration $t$, $\Mv w_t$ is obtained by 
\begin{equation} \label{eq:ridge}
\Mv w_t = \argmin{\Mv w} \left( \lambda \|\Mv w\|^2 + \|\Mv Z_t\Mv w - \Mv r_t\|^2 \right) \; \; \; ,
\end{equation}
where $\Mv Z_t = \begin{pmatrix}
\Mv z_{k_1,1},\cdots,\Mv z_{k_{t-1},t-1}
\end{pmatrix}^T$ contains the context vectors that correspond to the chosen actions as rows
and $\Mv r_t = 
\begin{pmatrix}
r_{k_1,1},\cdots, r_{k_{t-1},t-1}
\end{pmatrix}
$ contains the rewards revealed by the chosen actions as elements.

LinUCB runs an online procedure to solve
(\ref{eq:ridge}) and update~$\Mv w_t$. In particular,
LinUCB maintains a matrix $\Mv A_t = \Mv Z_t^T \Mv Z_t + \lambda \Mv I$ and a vector $\Mv b_t = \Mv Z_t^T \Mv r_t$ by
\begin{equation} \label{eq:ridge_Ab_update}
\begin{cases}
\,\Mv A_{t} =  \Mv A_{t-1} + \Mv z_{k_{t-1},t-1} \Mv z_{k_{t-1},t-1}^T \\
\,\Mv b_{t} = \Mv b_{t-1} + r_{k_{t-1},t-1} \Mv z_{k_{t-1},t-1} \\
\end{cases}
,
\end{equation}
where $\Mv A_0 = \lambda \Mv I$ and $\Mv b_0 =  \Mv 0$ are initialized
before the first iteration. Then, the solution to (\ref{eq:ridge}) is simply
\begin{equation} \label{eq:sol_ridge}
\Mv w_t = \Mv A_t^{-1} \Mv b_t \;\;\; .
\end{equation}

To maximize the cumulative reward, LinUCB uses the upper-confidence-bound technique to balance exploration and exploitation. 
That is, in each iteration $t$, LinUCB performs the action 
\begin{equation} \label{eq:upper_confidence_select}
k_t = \argmax{k} u_{k,t} \;\;\;,
\end{equation}
where 
\begin{equation}\label{eq:upper_confidence_value}
u_{k,t} = \Mv w_t^T\Mv z_{k,t} + \alpha \sqrt{\Mv z_{k,t}^T\Mv A_t^{-1}\Mv z_{k,t}} \;\;\; .
\end{equation}
The first term corresponds to the estimated reward of action~$k$ in iteration~$t$, and the second term represents the uncertainty of action $k$ under its context vector. The parameter $\alpha$ controls the preference between exploration (the second term) and exploitation (the first term).

We follow \cite{COMB}, a pioneer blending approach for active learning, to connect active learning with LSA and contextual bandit with LinUCB.
In particular, we treat each $\Mvt x_j \in \mathcal{D}_u$ as an action $k \in \{1, 2, \cdots, |\mathcal{D}_u|\}$. Then,
performing an action~$k_t$ in iteration $t$ by LinUCB is equivalent to querying the corresponding $\Mvt x_{k_t}$ by LSA.
The remaining issues are to specify what the context vectors $\Mvt z_{k, t}$ are and how the rewards~$r_{k_t, t}$ are calculated.
We first discuss our choice of the context vectors to achieve experience updating, and then illustrate our design
of the rewards, which represents active learning performance, in Section~\ref{sec:RF}.

As discussed in Section~\ref{sec:exp}, 
our active learning experience~$\Mv w$ is defined as the blending parameters of the set of scores $\bigl(s_1(\Mvt x_{k,t}, h_{t-1}),...,s_M(\Mvt x_{k,t}, h_{t-1})\bigr)$ given an unlabeled instance $\Mvt x_{k,t}$.
The definition allows a natural connection between LinUCB and LSA by setting
\begin{equation} \label{eq:def_z}
  \Mv z_{k, t} = \bigl(s_1(\Mvt x_{k,t}, h_{t-1}),...,s_M(\Mvt x_{k,t}, h_{t-1})\bigr).
\end{equation}
Then, the vector $\Mv w_t$ in LinUCB corresponds to the evolving experience $\Mv w$ calculated by ridge regression; the inner product~$\Mv w_t^T \Mv z_{k, t}$, which is the first term
of \eqref{eq:upper_confidence_value}, corresponds to the aggregated scoring function~$\hat{s}(\Mvt x_{k, t}, h_{t-1})$ that is made from both the current experience~$\Mv w_t$ and the human knowledge $\{s_m\}_{m=1}^M$. LSA queries an unlabeled instance with (\ref{eq:upper_confidence_select}) and \eqref{eq:upper_confidence_value},
which contains $\hat{s}(\cdot, \cdot)$ as well as an exploration term introduced by LinUCB, and updates the experience $\Mv w_t$ with (\ref{eq:sol_ridge}) and .


Recall that the goal of ridge regression within LinUCB is to provide a good estimate from the context vector to the reward.
We apply one trick in $\Mv z_{k, t}$ to improve the quality of the estimate. In particular, we add another element of $\Mv z_{k, t}[0]$, and set the element to a constant value of the previous reward
\begin{equation}
  \label{eq:def_z_fin}
  \Mv z_{k, t}[0] = r_{k_{t-1}, t-1} \; \; \; ,
\end{equation}
  where the rewards (including the edge case of $\Mv z_{k, 1}[0]$) will be defined in Section~\ref{sec:RF}. According to \eqref{eq:upper_confidence_value}, the added constant does not affect the choice of $k_t$, but it allows ridge regression to utilize the previous reward for estimating the current reward. In other words, the value provides a shared context on the active learning performance to assist the linear model. Empirically, we observe that the trick
indeed improves the quality of the estimate and the stability of LSA.



\begin{algorithm}[t]
\caption{Linear Strategy Aggregation}
\label{alg:LSA}
\begin{algorithmic}[1]

\setre{Parameters:}
\Require LinUCB balancing parameter $\alpha$, ridge regression parameter $\lambda$, minimum goodness parameter $\epsilon$, number of iterations $T$

\setre{Input:}
\Require labeled pool $\mathcal{D}_l$, unlabeled pool $\mathcal{D}_u$, scoring functions $\{s_1, s_2,\cdots,s_M\}$; a labeling oracle
\setre{Input:}
\setre{Begin:}
\Require
\State Initialize $\Mv A_0 = \lambda \Mv I, \Mv b_0 = \Mv 0$
	\For{$t=1,2,....,T$}
		\State Obtain contexts $\Mv z_{1,t}, \Mv z_{2,t}, ..., \Mv z_{|\mathcal{D}_u|,t}$ with (\ref{eq:def_z}) and (\ref{eq:def_z_fin})
		\State Obtain $u_{k_t,t}$, $\Mv z_{k_t,t}$ and $\td{\Mv x}_{k_t,t}$ with (\ref{eq:upper_confidence_select}) and (\ref{eq:upper_confidence_value})
		\State Query $\td{\Mv x}_{k_t}$ and get $\td{y}_{k_t}$ from the oracle
		\State Learn $h_t$ with $\mathcal{D}_l \cup \{(\Mvt x_{k_t},\td{y}_{k_t})\}$ 
		\State Obtain $v_t$ with (\ref{eq:reward_weight})
		\State Calculate $r_{k_t,t}$ with (\ref{eq:reward_sum})
		\State Update $\Mv A_t, \Mv b_t, \Mv w_t$ by $(\Mv z_{k_t, t}, r_{k_t, t})$ with (\ref{eq:ridge_Ab_update}) and (\ref{eq:sol_ridge})%
		\State $\mathcal{D}_l = \mathcal{D}_l \cup \{(\td{\Mv x}_{k_t},\td{y}_{k_t})\}$, $\mathcal{D}_u = \mathcal{D}_u \backslash \{\td{\Mv x}_{k_t}\}$
	\EndFor
\end{algorithmic}
\end{algorithm}

\subsection{Reward scheme} \label{sec:RF}

The only issue left for LSA is a properly designed reward
that represents active learning performance, or namely test accuracy in this work.
A state-of-the-art reward function proposed is called importance-weighted accuracy (IW-ACC), which is used in the \textit{Active Learning By Learning} (ALBL) approach~\cite{ALBL}.
IW-ACC weighs each instance in $\mathcal{D}_l$ with the inverse of the probability that the instance is queried,
and calculates the weighted accuracy as the reward.
The importance weighting allows IW-ACC to be an unbiased estimator of the test accuracy.

More specifically, in each iteration $t$ of ALBL, let $\Mvt x_{k_t}$ be the instance queried, $y_{k_t}$ be the obtained label, and $p_{k_t, t}$ be the probability of querying $\Mvt x_{k_t}$. Then, with
$v_t = p_{k_t, t}^{-1}$, 
IW-ACC is calculated as
\begin{equation}\label{eq:reward_sum}
  r_{k_\tau, \tau} = \frac{\sum_{t=1}^\tau v_t \llbracket h_\tau(\Mvt x_{k_t}) = y_{k_t} \rrbracket}{\sum_{t=1}^\tau v_t} \; \; \;,
\end{equation}
where $\llbracket \cdot \rrbracket$ is the indicator function.
The probability~$p_{k_t, t}$ reflects the \emph{goodness} of $\Mvt x_{k_t}$ in iteration $t$, and the
key idea of IW-ACC is to
assign $v_t$ as the inverse of the goodness to correct the sampling bias during active learning.

Nevertheless, unlike ALBL, LSA is a deterministic algorithm based on LinUCB. Thus, there is no $p_{k_t, t}$ and IW-ACC cannot be directly taken as the reward.
We thus propose a new reward scheme that mimics the key idea of IW-ACC. In our proposed scheme, each instance $\td{\Mv x}_{k_t}$ queried in iteration $t$ is weighted with 
\begin{equation}\label{eq:reward_weight}
v_t = \Bigl(\max(u_{k_t,t}, \epsilon)\Bigr) ^{-1}
\end{equation}
where $u_{k_t,t}$ is from (\ref{eq:upper_confidence_value}) and $\epsilon > 0$ is a small constant.

Recall that LSA maximizes over $u_{k,t}$ to decide the instance to be queried. That is, $u_{k, t}$ reflects the goodness of the unlabeled instance $\Mvt x_{k,t}$.
By using the inverse of $u_{k_t,t}$ as weights, our proposed scheme effectively meets the key idea of importance weighting behind IW-ACC while avoiding the need of probabilistic queries. The small constant $\epsilon > 0$ guards the rare edge cases of  $u_{k_t,t} \leq \epsilon$.

In the proposed LSA, the rewards are of another use of serving as $\Mv z_{k, t}[0] = r_{k_{t-1}, t-1}$ in \eqref{eq:def_z_fin}. When $t = 1$, there is technically no ``previous reward'' to use in \eqref{eq:def_z_fin}. The simplest choice would be taking $\Mv z_{k, 1}[0] = 0.5$ for representing the random-guessing accuracy. In this work, we heuristically take $\Mv z_{k, 1}[0]$ to be the training accuracy when learning from the initial $\mathcal{D}_l$ in order to provide a better shared context on the performance.

With the proposed scheme, the final piece of LSA is now complete.
In each iteration $t$, 
LSA simply runs LinUCB to query an unlabeled instance $\td{\Mv x}_{k_t}$ using (\ref{eq:upper_confidence_select}) and updates
the experience $\Mv w_t$ with (\ref{eq:sol_ridge}) by the context vector $\Mv z_{k_t,t}$ as well as the proposed reward $r_{k_t, t}$ in (\ref{eq:reward_sum}).
The details of LSA are listed in Algorithm~\ref{alg:LSA}.

\section{Active Learning Across Datasets} \label{Transfer_LSA}

LSA is now able to adaptively update the experience within any single dataset.
Our next goal is to achieve experience transfer across datasets,
with the hope of improving active learning performance.
We thus design an extension of LSA, called \textit{Transfer LSA} (T-LSA), that takes the learned experience as a reference when conducting active learning on the current dataset.

Our design is motivated from an earlier work that focuses on personalized handwriting recognition \cite{bias_reg}. 
The main idea of the work is to first learn a generic handwriting recognizer $\Mv w_{\Mt{gen}}$ 
by SVM from a large amount of handwriting data of all people.
The personalized handwriting recognizer $\Mv w$ is then learned from a small amount of individual data
via a \textit{Biased Regularization SVM} (BRSVM).
BRSVM replaces the $\ell_2$ regularization term
$\frac{1}{2}\|\Mv w\|^2$ in the objective function of SVM 
with a biased regularization term $\frac{1}{2}\|\Mv w - \Mv w_{\Mt{gen}}\|^2$
to enforce the personalized $\Mv w$ to be close to the generic $\Mv w_{\Mt{gen}}$.

BRSVM for personalized handwriting recognizer allows learning of $\Mv w$ with the prior knowledge of $\Mv w_{\Mt{gen}}$ as a reference point.
In our cross-dataset active learning problem, we intend to take $\Mv w_{\Mt{prev}}$, the experience learned from other datasets,
as our reference point. 
For simplicity, let us first assume that
$\Mv w_{\Mt{prev}}$ comes from the experience of active learning from one previous dataset.
That is, $\Mv w_{\Mt{prev}} = \Mv w_{T}$ learned from $(\mathcal{D}_l^{(1)}, \mathcal{D}_u^{(1)})$.
Recall that $\Mv w_t$ in LSA is the ridge-regression solution of (\ref{eq:ridge}). Then,
we borrow the idea of BRSVM to replace $\frac{1}{2}\|\Mv w\|^2$ with $\frac{1}{2}\|\Mv w - \Mv w_{\Mt{prev}}\|^2$ as our regularization term.
That is, biased regularization can be simply achieved by solving
\begin{equation} \label{eq:bias_ridge}
\Mvh w_t = \argmin{\Mv w} \lambda \|\Mv w - \Mv w_{\Mt{prev}}\|^2 + \|\Mv Z_t\Mv w - \Mv r_t\|^2
\end{equation}
instead. The close-form solution is
\begin{equation}\label{eq:sol_bias_ridge}
\Mvh w_t = (\Mv Z_t^T \Mv Z_t+\lambda \Mv I)^{-1} (\Mv Z_t^T \Mv r_t + \lambda \Mv w_{\Mt{prev}})
\end{equation}
The parameter $\lambda$ now represents the trust of previous experience.

To integrate (\ref{eq:sol_bias_ridge}) into LSA, we need to update $\Mvh w_t$ online like \eqref{eq:ridge_Ab_update} and \eqref{eq:sol_ridge}.
  Recall that \eqref{eq:ridge_Ab_update} maintains $\Mv A_t = \Mv Z_t^T \Mv Z_t + \lambda \Mv I$ and $\Mv b_t = \Mv Z_t^T \Mv r_t$. Then,
  \eqref{eq:sol_bias_ridge} can be re-written as
\begin{equation} \label{eq:sol_bias_ridge_online}
\hat{\Mv w}_t = \Mv A_t^{-1}(\underbrace{\Mv b_t+\lambda \Mv w_{\Mt{prev}}}_{\Mv b_t'}).
\end{equation} 

Notice that the only difference between (\ref{eq:sol_ridge}) and (\ref{eq:sol_bias_ridge_online}) 
is the term $\lambda \Mv w_{\Mt{prev}}$ between $\Mv b_t$ and $\Mv b_t'$.  
Thus, we can easily achieve biased regularization in T-LSA by replacing $\Mv b_0=\Mv 0$ in LSA with $\Mv b_0'=\lambda \Mv w_{\Mt{prev}}$ and maintaining $\Mv b_t'$ instead of $\Mv b_t$. The weight vector $\hat{\Mv w}_t$ can then be updated
online with $\Mv A_t^{-1} \Mv b_t'$ in~\eqref{eq:sol_bias_ridge_online}. When $\Mv w_{\Mt{prev}} = \Mv 0$, which means zero experience, biased regularization falls back to usual $\ell_2$ regularization and T-LSA falls back to LSA.

We now consider the full setup of cross-dataset active learning, as defined in Section~\ref{background},
where a sequence of datasets, $(\Mc{D}^{(1)}_l, \Mc D^{(1)}_u),\cdots,(\Mc D^{(Q)}_l, \Mc D^{(Q)}_u)$, is presented. 
Let $\Mvh w^{(1)}$ be the experience learned from $(\Mc D^{(1)}_l, \Mc D^{(1)}_u)$.
When learning $\Mv w_t$ on~$(\Mc{D}^{(2)}_l, \Mc D^{(2)}_u)$ using $\Mv w_{\Mt{prev}} = \Mvh w^{(1)}$ as the reference point
in \eqref{eq:bias_ridge}, the first term $\lambda \|\Mv w - \Mv w_{\Mt{prev}}\|^2$ allows the information of the earlier experience to be somewhat preserved, and the second term $\|\Mv Z_t\Mv w - \Mv r_t\|^2$ allows new experience to be accumulated. Thus,~$\Mvh w^{(2)}$ learned from
$(\Mc D^{(2)}_l, \Mc D^{(2)}_u)$ contains experience from both the first and the second datasets. It is then natural to learn
$\Mvh w^{(3)}$ on $(\Mc D^{(3)}_l, \Mc D^{(3)}_u)$ with $\Mv w_{\Mt{prev}} = \Mvh w^{(2)}$, or more generally learn
$\Mvh w^{(q)}$ on $(\Mc D^{(q)}_l, \Mc D^{(q)}_u)$ with $\Mv w_{\Mt{prev}} = \Mvh w^{(q-1)}$ for $q = 2,\cdots,Q$.
The simple use of $\Mv w_{\Mt{prev}} = \Mvh w^{(q-1)}$ completes the design of the full T-LSA algorithm, as
listed in Algorithm~\ref{alg:Transfer_LSA}. For simplicity, we overload $\Mv b_t$ to denote $\Mv b_t'$ in Algorithm~\ref{alg:Transfer_LSA}.


With the help of biased regularization, T-LSA achieves cross-dataset active learning.
When the experience is helpful, which possibly happens when transferring experience from more related datasets, T-LSA utilizes the experience
to speedup exploration in the wild. When the experience is not so helpful, which can mean a negative transfer in the terminology of transfer learning, the second term $\|\Mv Z_t\Mv w - \Mv r_t\|^2$ in \eqref{eq:bias_ridge} allows new experience to be adaptively accumulated.
In Section \ref{sec:exp_cross_datasets}, we will empirically study how different kinds of experience affect the performance of T-LSA.

\begin{algorithm}[t]
\caption{Transfer LSA}
\label{alg:Transfer_LSA}
\begin{algorithmic}[1]

\setre{Parameters:}
\Require Same as parameters for Algorithm~\ref{alg:LSA}

\setre{Input:}
\Require Datasets sequence $(\mathcal{D}_l^{(1)}, \mathcal{D}_u^{(1)})$,...,$(\mathcal{D}_l^{(Q)}, \mathcal{D}_u^{(Q)})$, scoring functions for Algorithm~\ref{alg:LSA}
\setre{Input:}
\setre{Begin:}
\Require
        \State $\Mv w_{\Mt{prev}} \gets \Mv 0$
    \For{$q=1,2,\cdots,Q$}
		\State Initialize Algorithm~\ref{alg:LSA} (LSA) with $(\Mv A_0, \Mv b_0) = (\lambda \Mv I, \lambda \Mv w_{\Mt{prev}})$ instead
		\State Run the initialized LSA on $(\mathcal{D}_l^{(q)}, \mathcal{D}_u^{(q)})$ and obtain experience $\Mvh w^{(q)}$
		\State $\Mv w_{\Mt{prev}} \gets \Mvh w^{(q)}$
	\EndFor
\end{algorithmic}
\end{algorithm}

\section{Experiment} \label{experiment}
We couple the following key active learning algorithms with our proposed approaches, LSA and T-LSA, to validate their empirical performance. 
The algorithms, as illustrated in Section~\ref{background}, are
\begin{enumerate}
\item UNCERTAIN: uncertainty sampling with SVM \cite{uncertainty_svm}.
\item REPRESENT: representative sampling based on $k$-mean clustering \cite{representative}. Because the uncertainty part is essentially the same as UNCERTAIN, we only take the scoring function for representativeness for blending.
\item DUAL: another representative sampling approach using mixture-of-Gaussian weighted uncertainty as scoring function \cite{DUAL}.
\item QUIRE: another representative sampling approach using the min-max view of label-assignment to optimize the scoring function \cite{QUIRE}.
\end{enumerate}
We take logistic regression as our base classification model, and use the $\ell_2$-regularized logistic regression solver of LIBLINEAR \cite{liblinear} with default parameters to learn a classifier from the model.

We conduct experiments on two sets of benchmark datasets.
The first set is commonly used to validate pool-based active learning approaches for binary classification, and is taken to validate not only the competitiveness of LSA versus other approaches, but also to examine the potential of T-LSA for cross-dataset active learning with heterogeneous datasets.
The first benchmark set include the following eight datasets from the UCI repository \cite{UCI}:
\textit{austra}, \textit{breast}, \textit{diabetes}, \textit{german}, 
\textit{heart},  \textit{letterMvsN}, \textit{liver}, and \textit{wdbc}, 
where the dataset \textit{letterMvsN} is constructed from a multi-class dataset \textit{letter}.

The second set, which contains two datasets of handwritten digit recognition, \textit{USPS} and \textit{MNIST}, is used in several previous studies of multi-task learning~\cite{mtl_1, mtl_2}. We take the second set to examine the potential of T-LSA for cross-dataset active learning with homogeneous datasets.
We follow~\cite{mtl_1} to reduce the feature dimensions of \textit{USPS} and \textit{MNIST} to $87$ and $62$ respectively with principal component analysis.

For the larger datasets \textit{letterMvsN}, \textit{USPS} and \textit{MNIST}, we randomly keep only $2000$ examples to make experiments sufficiently efficient. Then,
we split each dataset randomly with $50\%$ for training and $50\%$ for testing, 
We take the training set as our unlabeled pool $\mathcal{D}_u$, and the test
set for reporting active learning performance. 
We randomly select~$4$ instances from the unlabeled pool $\mathcal{D}_u$ as our initial labeled pool $\mathcal{D}_l$. Experiments on each dataset are averaged over $10$ times.

\begin{figure*}[t]
\begin{center}
\centering
    \begin{subfigure}[b]{0.25\textwidth}
                \includegraphics[width=\textwidth]{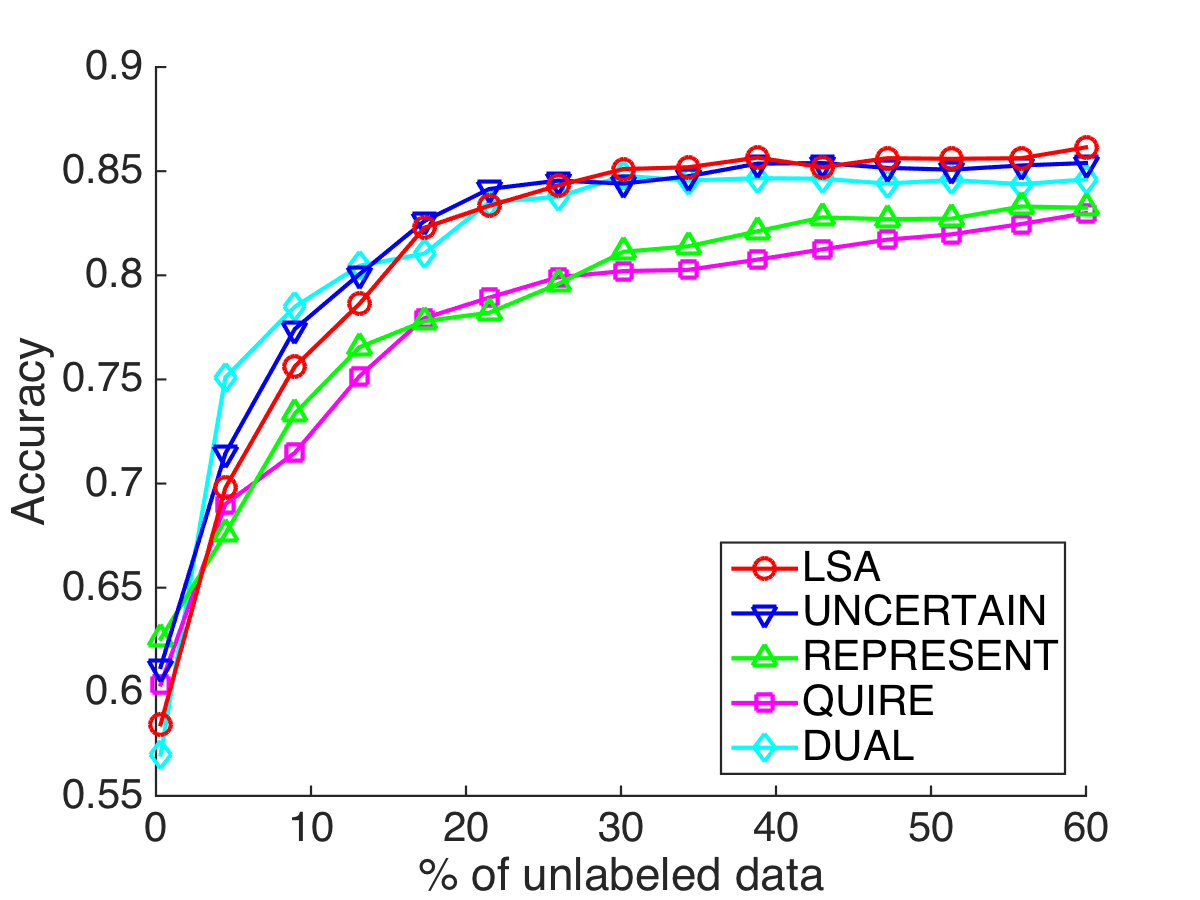}
                \caption{\textit{austra}}
                \label{fig:austra_lsa_underlying}
        \end{subfigure}%
    \begin{subfigure}[b]{0.25\textwidth}
                \includegraphics[width=\textwidth]{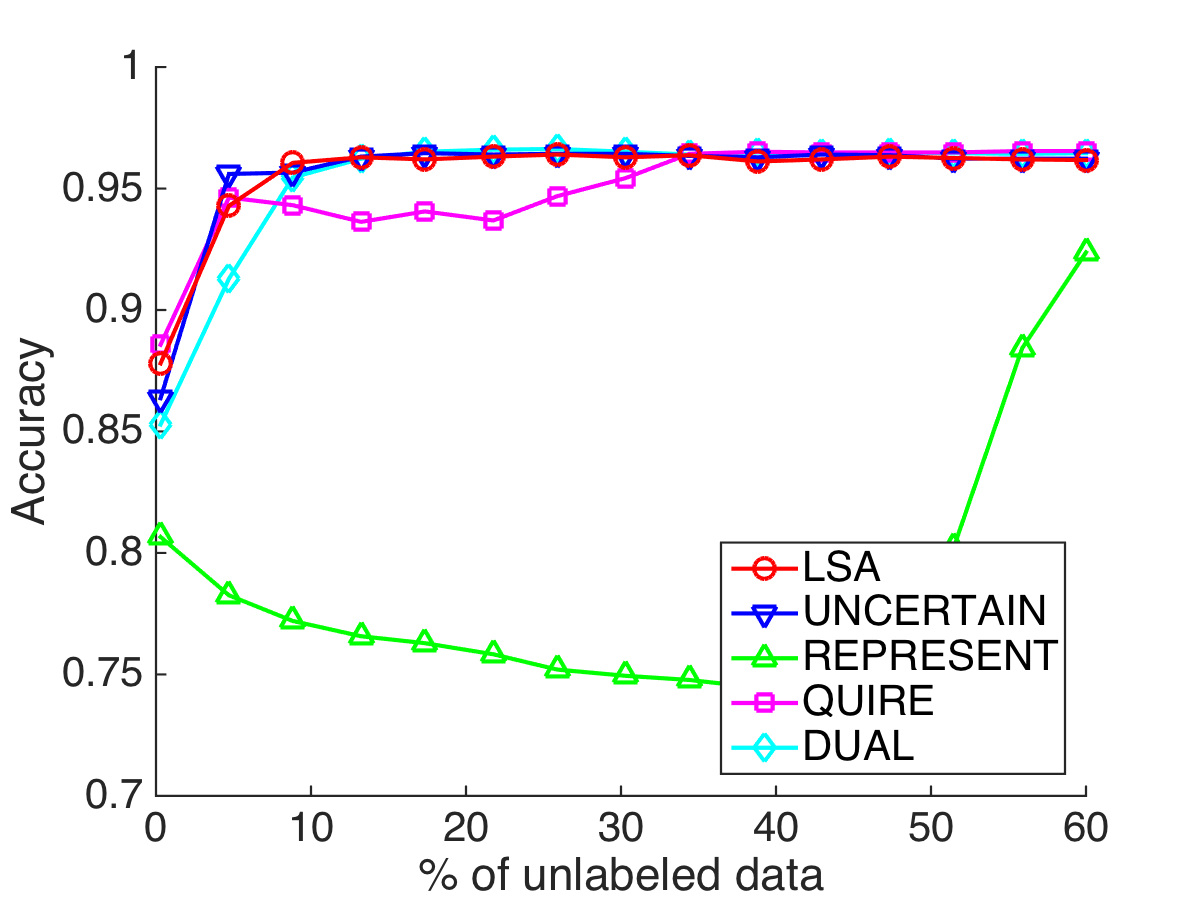}
                \caption{\textit{breast}}
                \label{fig:breast_lsa_underlying}
        \end{subfigure}%
    \begin{subfigure}[b]{0.25\textwidth}
                \includegraphics[width=\textwidth]{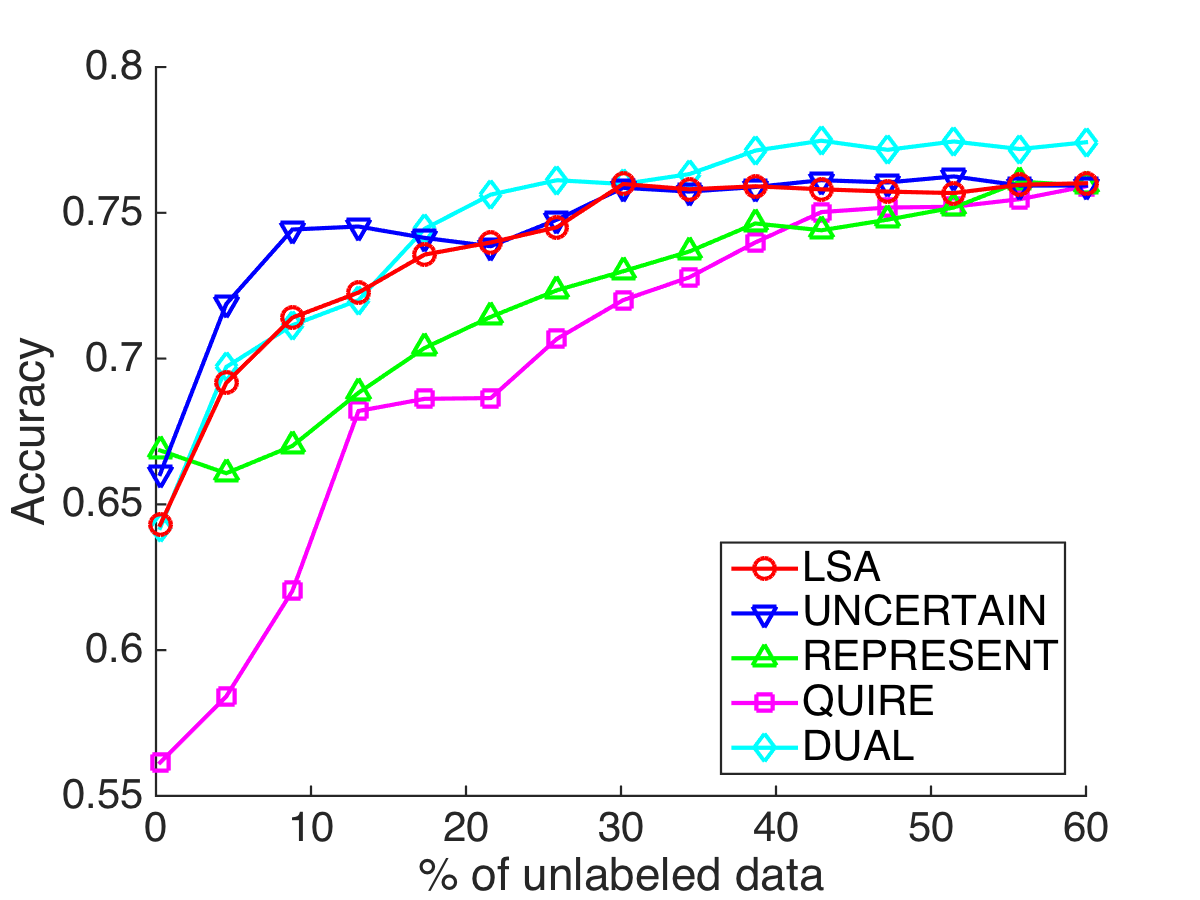}
                \caption{\textit{diabetes}}
                \label{fig:diabetes_lsa_underlying}
        \end{subfigure}%
    \begin{subfigure}[b]{0.25\textwidth}
                \includegraphics[width=\textwidth]{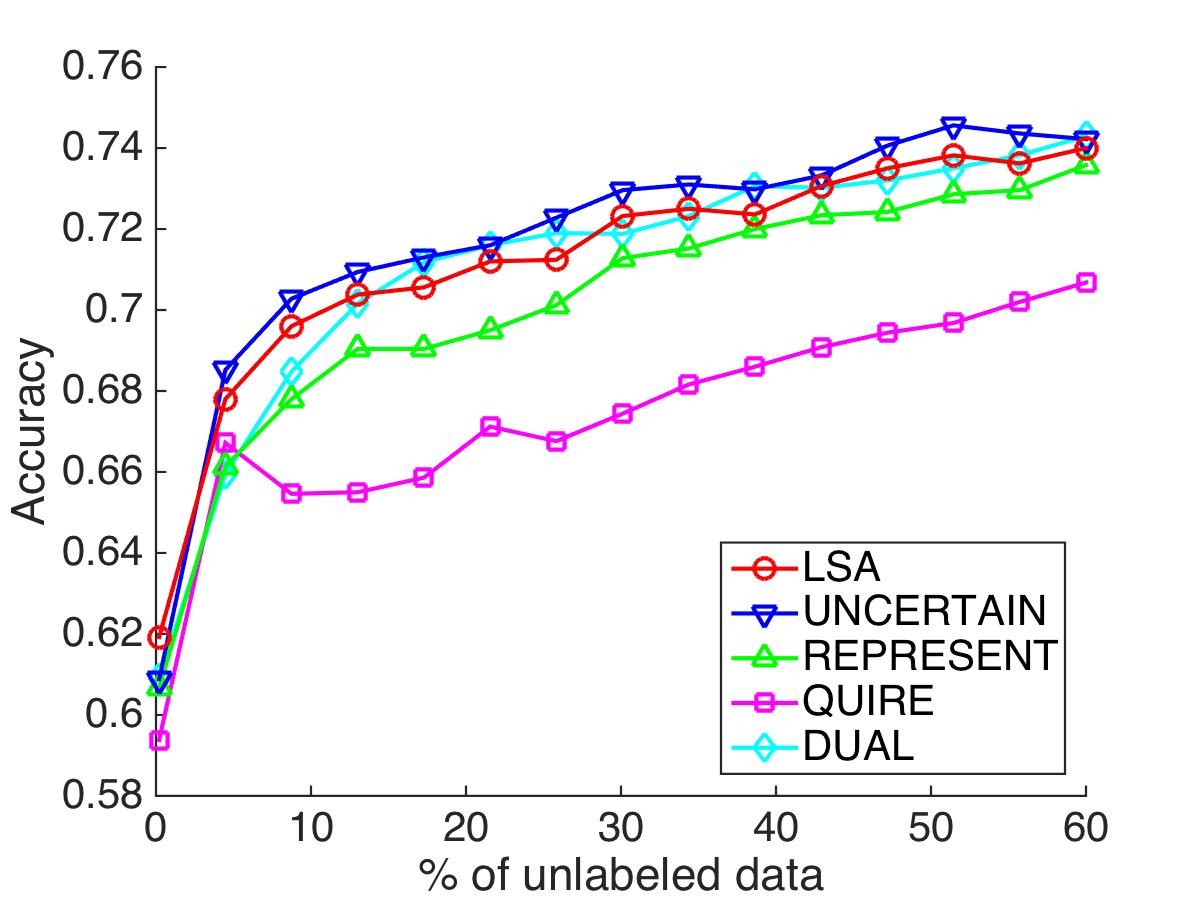}
                \caption{\textit{german}}
                \label{fig:german_lsa_underlying}
    \end{subfigure}%
    
    \begin{subfigure}[b]{0.25\textwidth}
                \includegraphics[width=\textwidth]{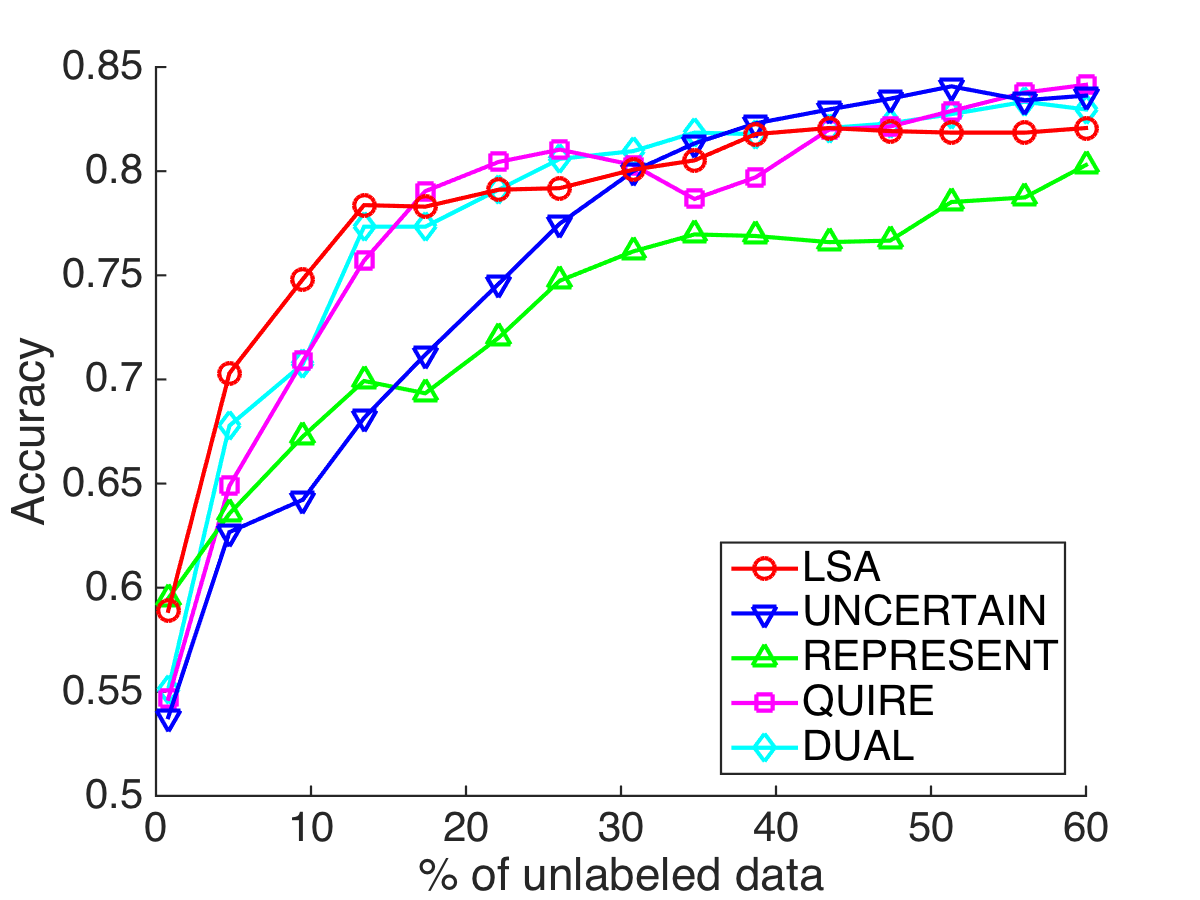}
                \caption{\textit{heart}}
                \label{fig:heart_lsa_underlying}
        \end{subfigure}%
    \begin{subfigure}[b]{0.25\textwidth}
                \includegraphics[width=\textwidth]{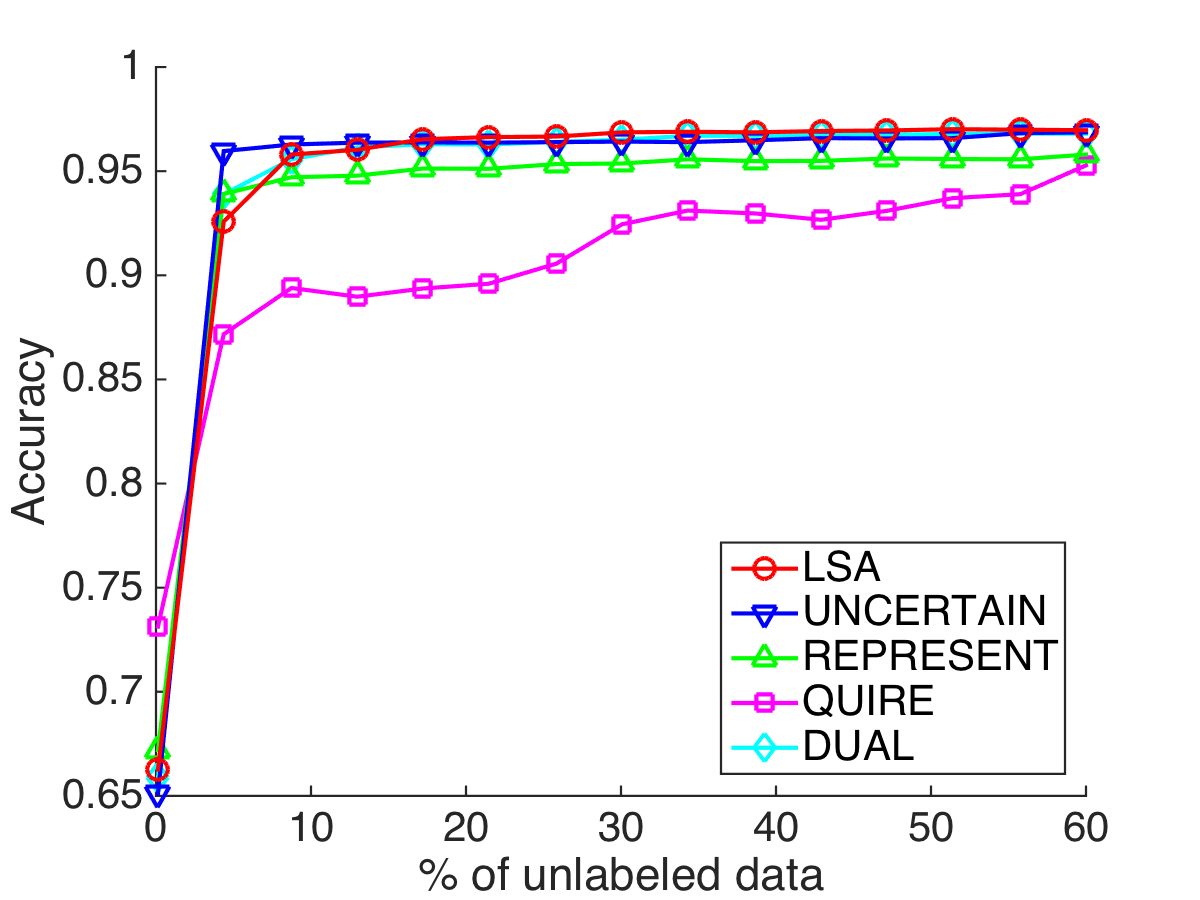}
                \caption{\textit{letterMvsN}}
                \label{fig:letter_MvsN_lsa_underlying}
        \end{subfigure}%
    \begin{subfigure}[b]{0.25\textwidth}
                \includegraphics[width=\textwidth]{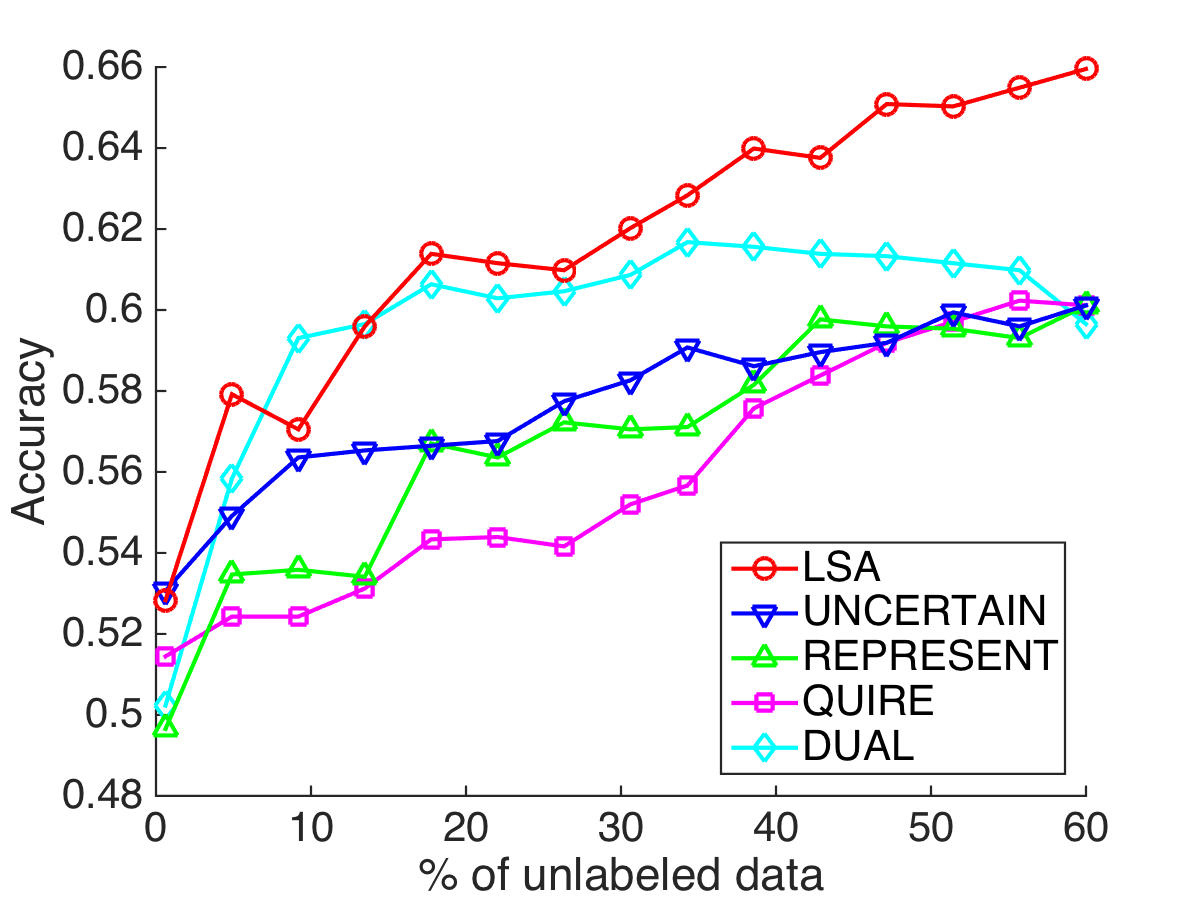}
                \caption{\textit{liver}}
                \label{fig:liver_lsa_underlying}
        \end{subfigure}%
    \begin{subfigure}[b]{0.25\textwidth}
                \includegraphics[width=\textwidth]{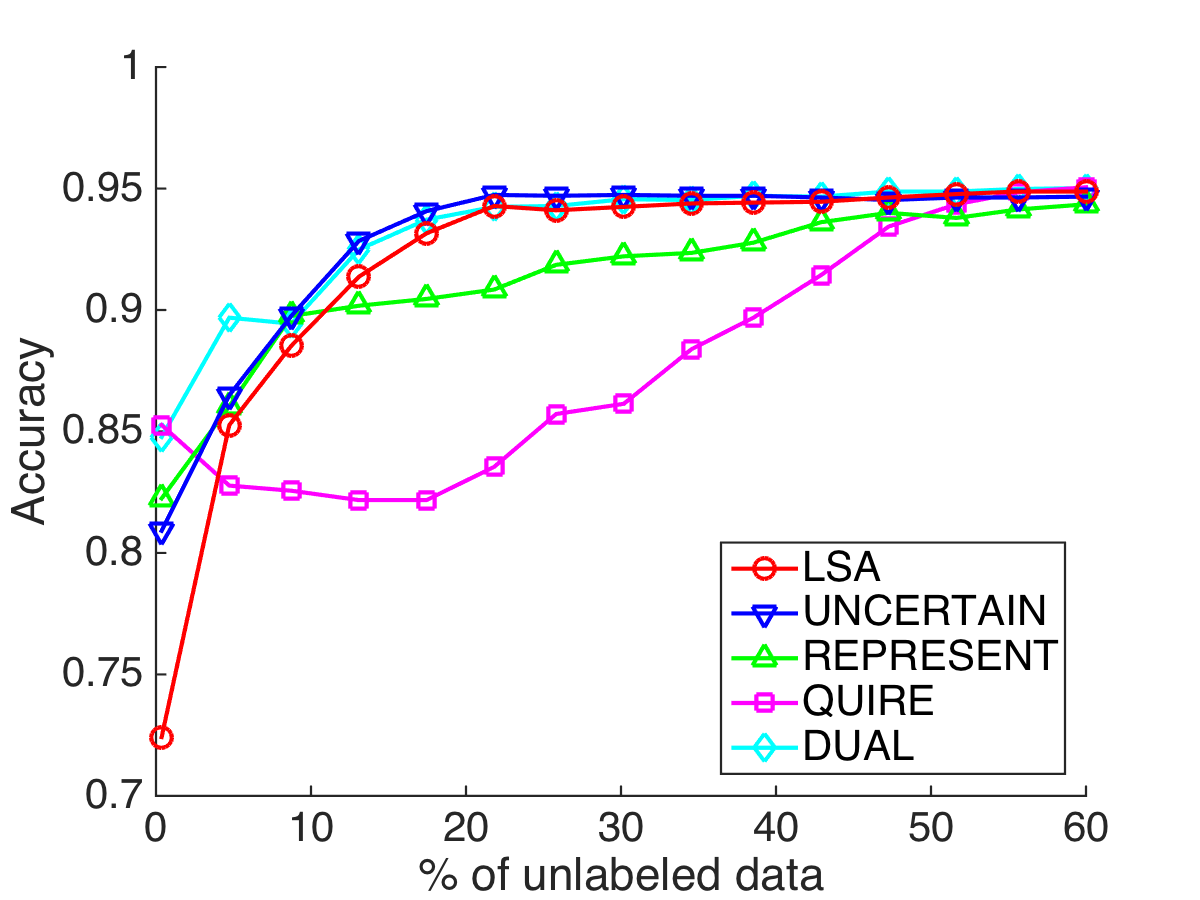}
                \caption{\textit{wdbc}}
                \label{fig:wdbc_lsa_underlying}
        \end{subfigure}%
\end{center}
\caption{Test Accuracy of LSA versus underlying strategies}
\label{fig:lsa_underlying}
\end{figure*}

\begin{figure*}[t]
\begin{center}
\centering
    \begin{subfigure}[b]{0.25\textwidth}
                \includegraphics[width=\textwidth]{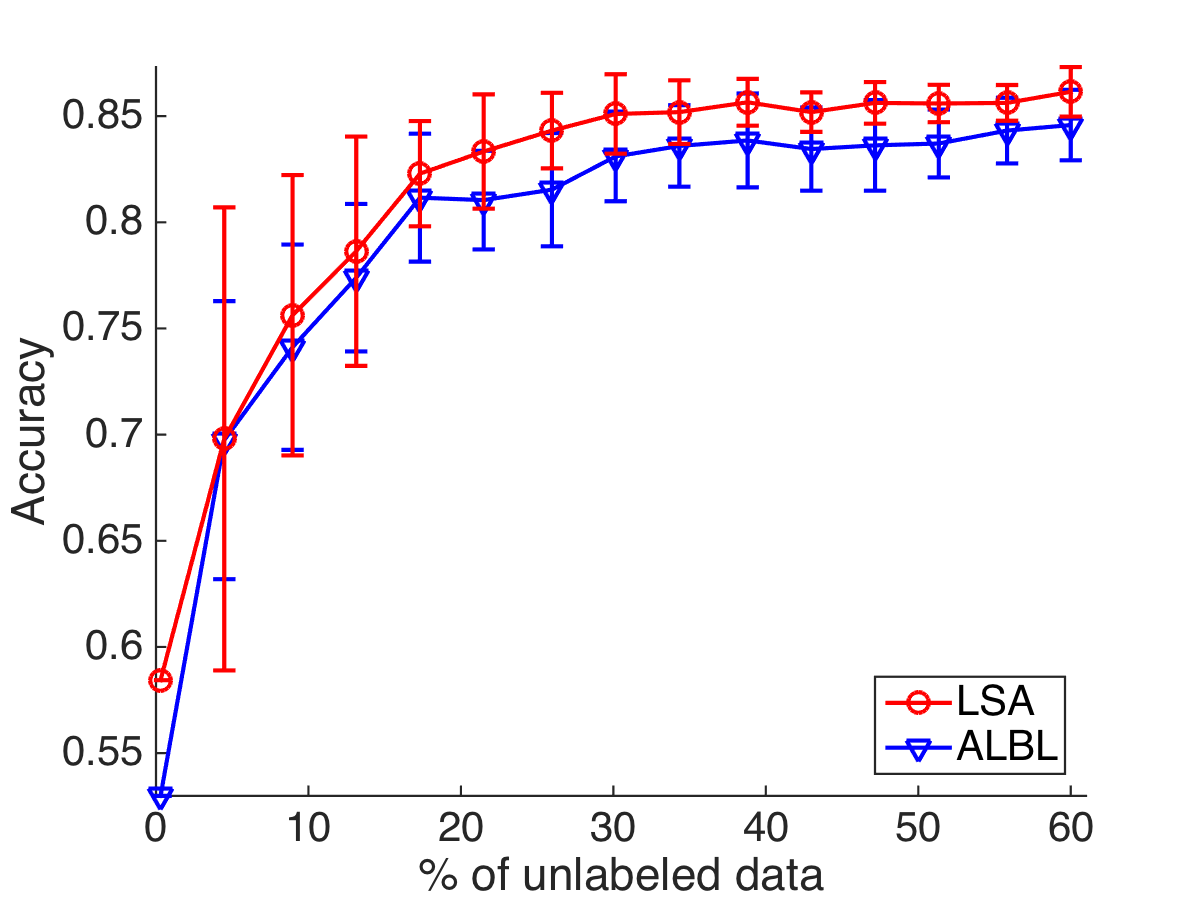}
                \caption{\textit{austra}}
                \label{fig:austra_lsa_albl}
        \end{subfigure}%
    \begin{subfigure}[b]{0.25\textwidth}
                \includegraphics[width=\textwidth]{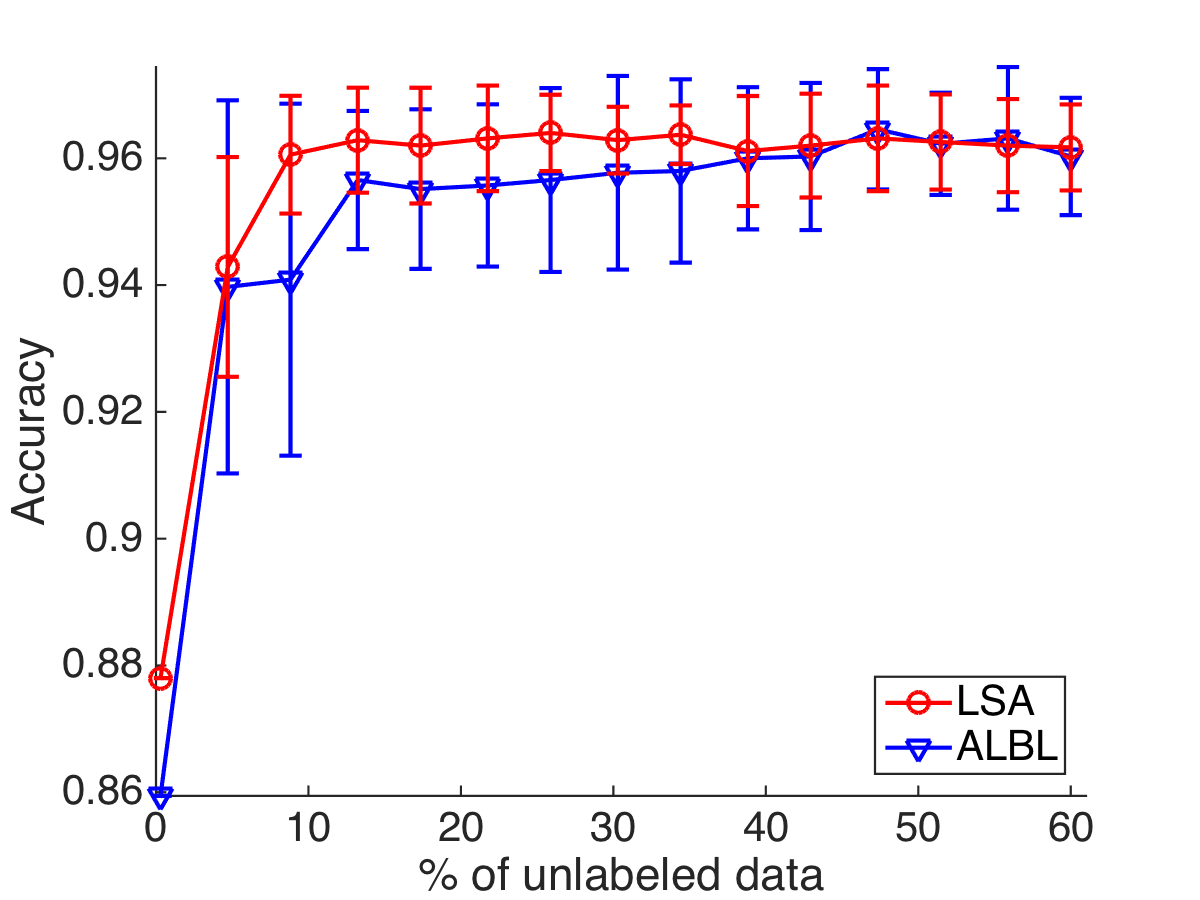}
                \caption{\textit{breast}}
                \label{fig:breast_lsa_albl}
        \end{subfigure}%
    \begin{subfigure}[b]{0.25\textwidth}
                \includegraphics[width=\textwidth]{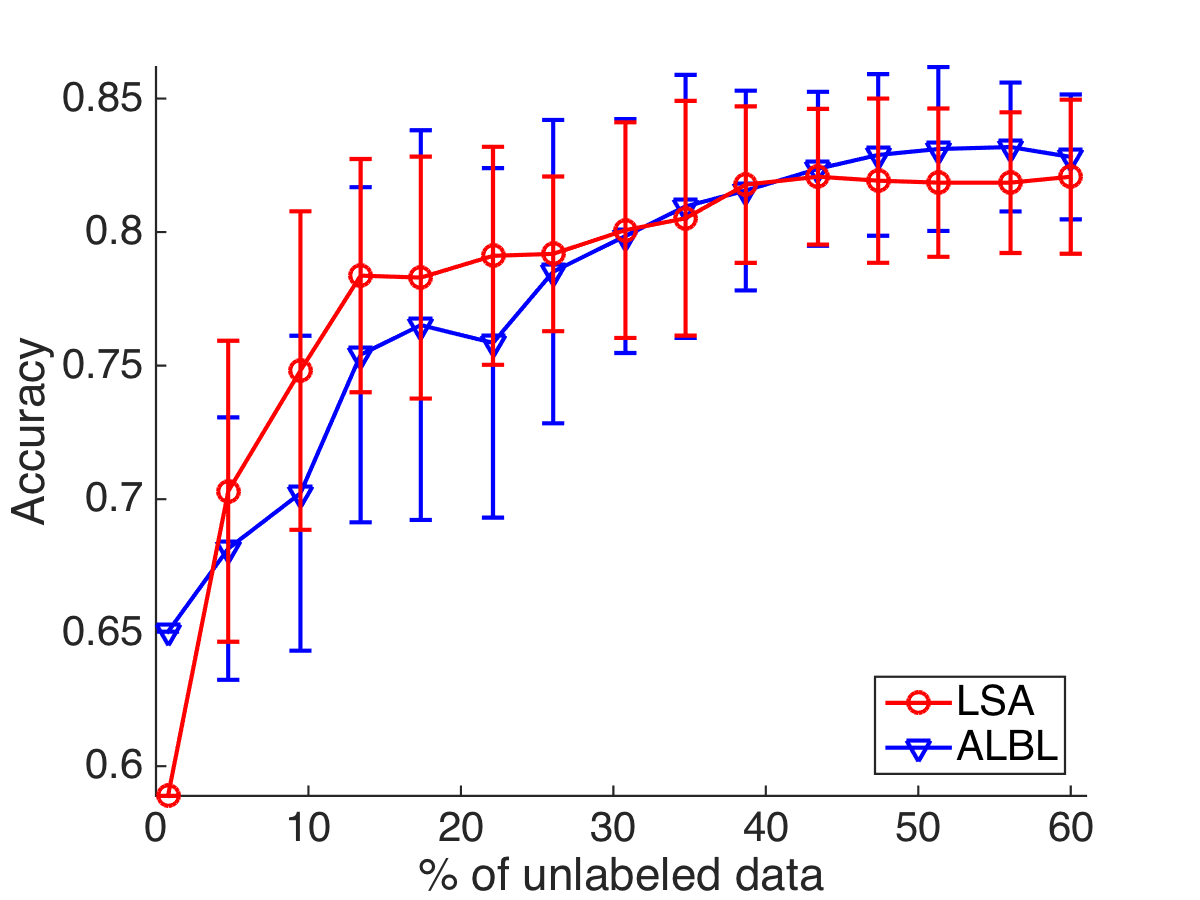}
                \caption{\textit{heart}}
                \label{fig:heart_lsa_albl}
        \end{subfigure}%
    \begin{subfigure}[b]{0.25\textwidth}
                \includegraphics[width=\textwidth]{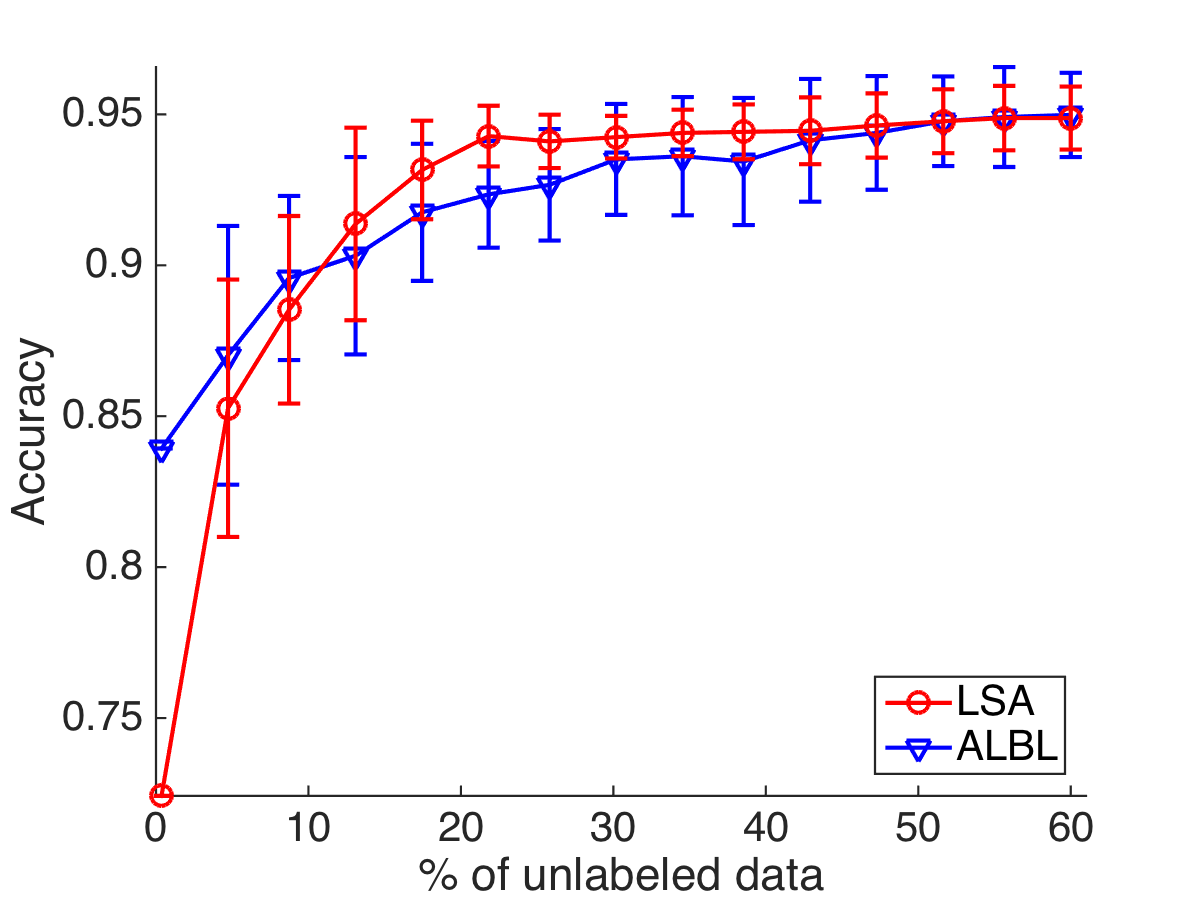}
                \caption{\textit{wdbc}}
                \label{fig:wdbc_lsa_albl}
        \end{subfigure}%
\end{center}
\caption{Test Accuracy of LSA versus ALBL}
\label{fig:lsa_albl}
\end{figure*}





\begin{table}[t]
\caption{LSA versus underlying algorithms based on $t$-test at 90\% confidence level (\#win/\#tie/\#loss)}
\label{tbl:lsa_underlying}
\begin{center}
\resizebox{0.48\textwidth}{!}{%
\begin{tabular}{l c c c c c c c l}
\hline 
\multirow{2}{*}{\bf{rank}} & \multicolumn{7}{c}{\bf percentage of queried instances} & \multirow{2}{*}{total}\\
\cline{2-8}
& 5\% & 10\% & 15\% & 20\% & 30\% & 40\% & 50\% &\\
\hline \hline 
1st& 0/6/2 & 0/7/1 & 0/7/1 & 0/8/0 & 0/8/0 & 0/8/0 & 1/6/1 & 1/50/5 \\
2nd& 0/8/0 & 0/8/0 & 1/7/0 & 0/8/0 & 0/8/0 & 1/7/0 & 1/7/0 & 3/53/0 \\
3rd& 1/7/0 & 4/4/0 & 4/4/0 & 6/2/0 & 5/3/0 & 4/4/0 & 3/5/0 & 27/29/0 \\
4th& 4/4/0 & 7/1/0 & 8/0/0 & 8/0/0 & 8/0/0 & 7/1/0 & 6/2/0 & 48/8/0 \\
\hline
 total& 5/25/2 & 11/20/1 & 13/18/1 & 14/18/0 & 13/19/0 & 12/20/0 & 11/20/1 & 79/140/5 \\
\hline 
\end{tabular}
}
\end{center}
\end{table}

\begin{table}[t]
\caption{LSA versus ALBL based on $t$-test at 90\% confidence level (\#win/\#tie/\#loss)}
\label{tbl:lsa_albl}
\begin{center}
\resizebox{0.48\textwidth}{!}{%
\begin{tabular}{l c c c c c c c l}
\hline 
& \multicolumn{7}{c}{\bf percentage of queried instances} & \multirow{2}{*}{total}\\
\cline{2-8}
& 5\% & 10\% & 15\% & 20\% & 30\% & 40\% & 50\% &\\
\hline \hline 
ALBL& 0/8/0 & 2/6/0 & 2/6/0 & 2/6/0 & 2/6/0 & 2/5/1 & 3/5/0 & 13/42/1 \\
\hline 
\end{tabular}
}
\end{center}
\end{table}

We will first compare LSA with the four underlying active learning algorithms and the state-of-the-art ALBL approach~\cite{ALBL} for blending those algorithms on single datasets. Then, we will compare T-LSA with LSA and ALBL under the cross-dataset setting to understand the effectiveness of experience transfer. For fairness, we will also na{\"i}vely extend ALBL to T-ALBL as illustrated in Section~\ref{background}, and take T-ALBL for comparison. In particular, T-ALBL
initializes the internal probability distribution with the previously learned distribution to achieve experience transfer.


Parameter tuning of active learning is known to be hard~\cite{ALBL}. In our experiments, we run the approaches
on several parameter combinations, and report the result of the best combination. Practically, existing blending approaches like ALBL~\cite{ALBL} or COMB \cite{COMB}
can then be run on top of the combinations to adaptively approximate the best result.
Specifically, for the experiments on single datasets, we run LSA with $\lambda = 1$ and $\alpha \in \{1.5,2.0,2.5\}$. 
For the experiments of cross-dataset active learning, we fix $\alpha = 1.5$ and run LSA and T-LSA with $\lambda = 1$ and $\lambda \in \{1,5,10\}$ respectively.
For the parameters of other algorithms, we follow the recommended parameters provided in the paper/codes from the authors.

We do not include another adaptive blending approach of COMB \cite{COMB} for two reasons:
\begin{enumerate}
\item ALBL is known to outperform COMB on single datasets~\cite{ALBL}.
\item Unlike ALBL, which maintains an internal probabilistic distribution on the active learning algorithms, COMB maintains the distribution on the unlabeled instances. It is non-trivial to transfer the distribution as experience to other
  datasets with different number of instances.
\end{enumerate}

\subsection{Experiments on Single Datasets}\label{sec:exp_single}
We first compare LSA with the four underlying active learning algorithms on the first set of eight benchmark datasets, and plot the test accuracy under different percentages of queries in Fig. \ref{fig:lsa_underlying}. 
From the results, we can observe that LSA is usually close to the best curves of the four algorithms after querying $10\%$ of unlabeled instances. The results demonstrate that LSA is effective in terms of blending human knowledge towards
decent query decisions. 
The less-strong performance of LSA in the first $10\%$ of queries
hints the need of using experience to guide exploration instead of starting from zero experience.

The results in Fig.~\ref{fig:lsa_underlying} is further supported by Table~\ref{tbl:lsa_underlying} with~$t$-tests at $90\%$ significance level. The tests compare LSA with the underlying algorithms at different ranks.
Table~\ref{tbl:lsa_underlying} indicates that LSA often yields competitive performance with the best underlying algorithm, and is always no-worse than the second best. The results in Fig.~\ref{fig:lsa_underlying} and Table~\ref{tbl:lsa_underlying} confirm that LSA to be a decent adaptive blending approach for
active learning, just like its ancestors of ALBL~\cite{ALBL} and COMB~\cite{COMB}. Note that LSA is a deterministic approach while ALBL and COMB are both probabilistic.

\begin{figure*}[t]
\begin{center}
\centering
    \begin{subfigure}[b]{0.25\textwidth}
                \includegraphics[width=\textwidth]{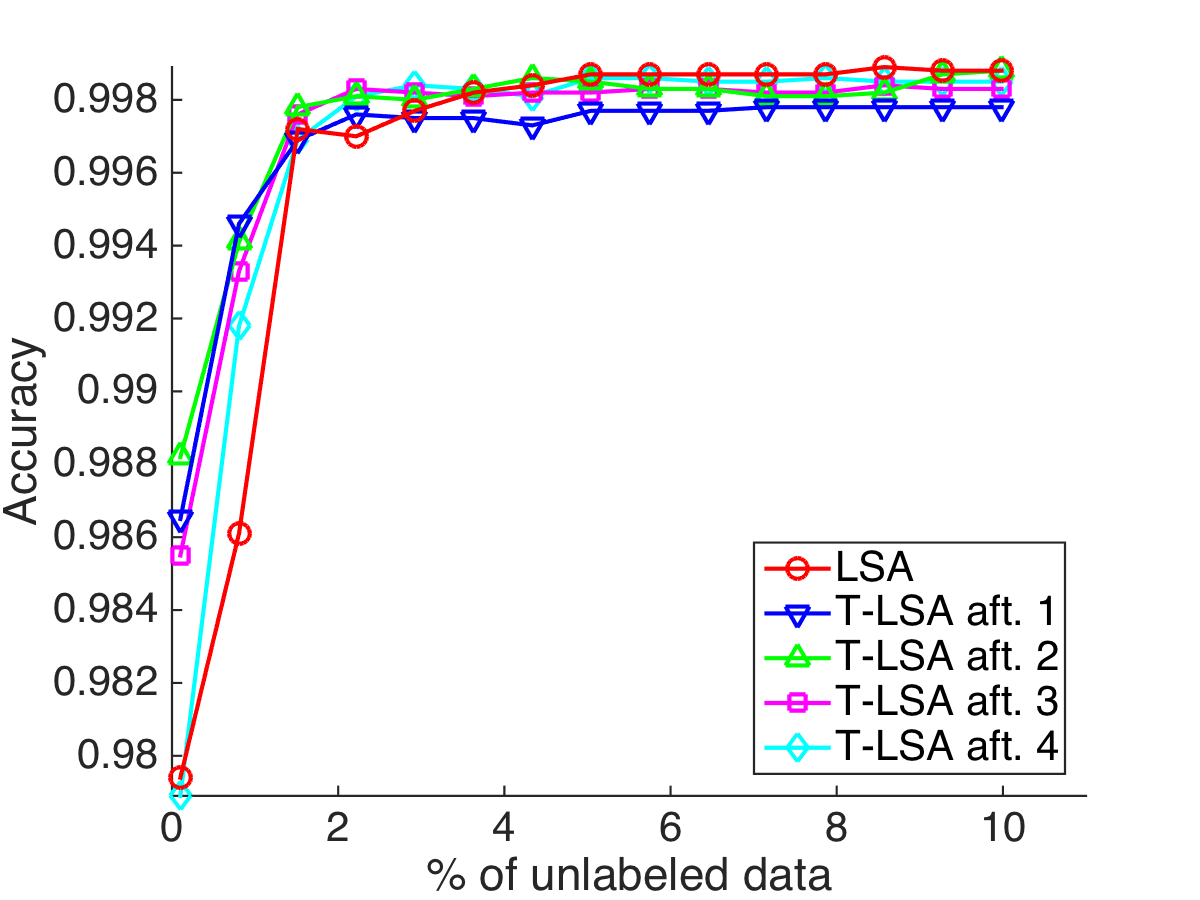}
                \caption{\textit{USPS 0vs1}}
                \label{fig:USPS_0vs1_lsa_tlsa_USPS}
        \end{subfigure}%
    \begin{subfigure}[b]{0.25\textwidth}
                \includegraphics[width=\textwidth]{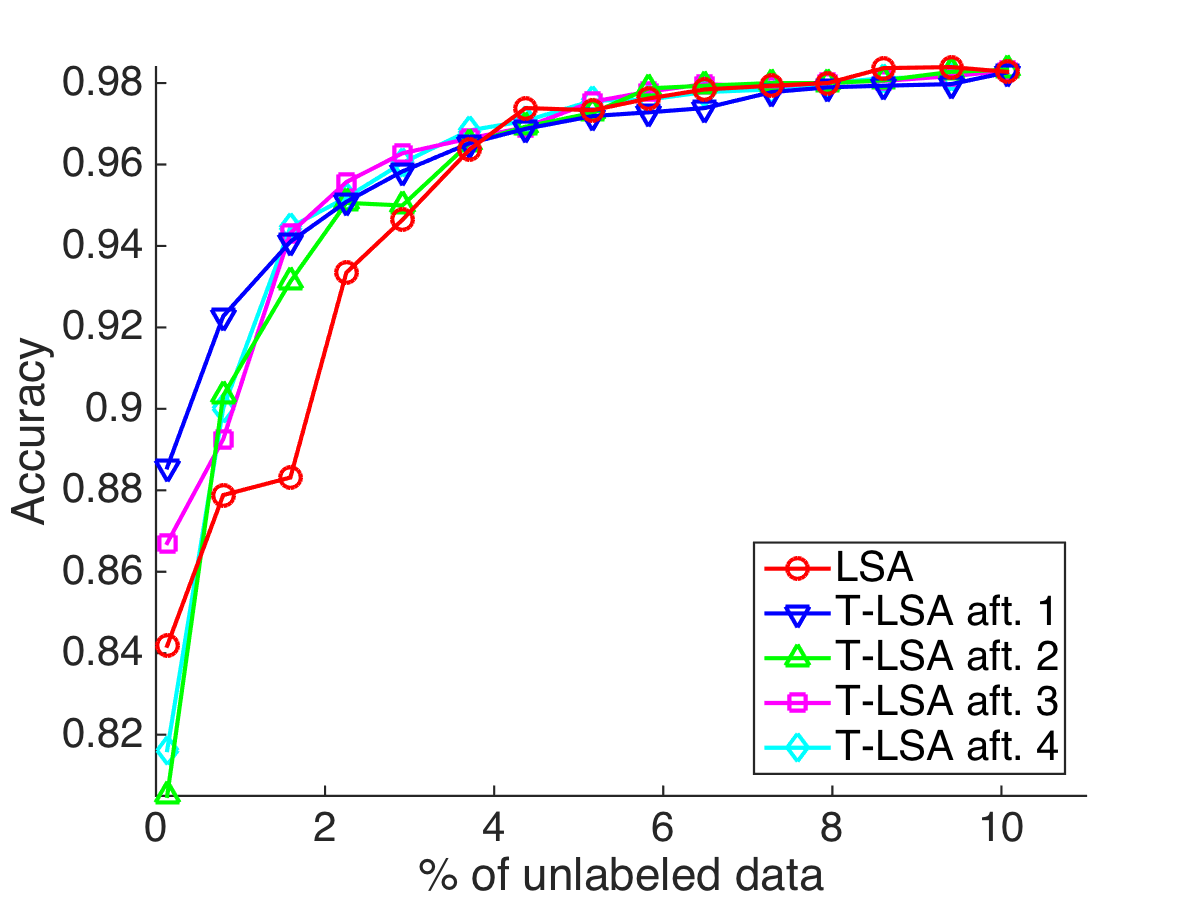}
                \caption{\textit{USPS 8vs9}}
                \label{fig:USPS_8vs9_lsa_tlsa_USPS}
        \end{subfigure}%
    \begin{subfigure}[b]{0.25\textwidth}
                \includegraphics[width=\textwidth]{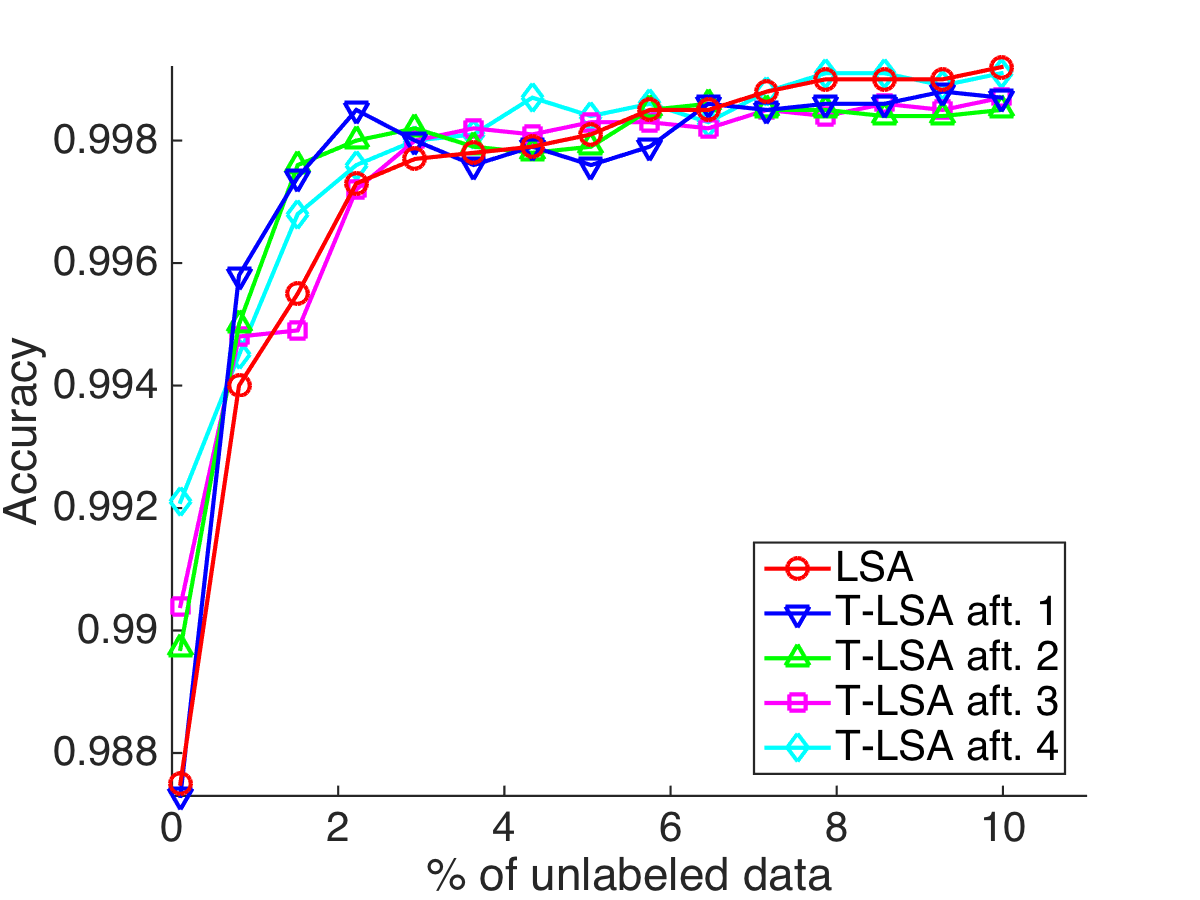}
                \caption{\textit{MNIST 0vs1}}
                \label{fig:MNIST_0vs1_lsa_tlsa_MNIST}
        \end{subfigure}%
    \begin{subfigure}[b]{0.25\textwidth}
                \includegraphics[width=\textwidth]{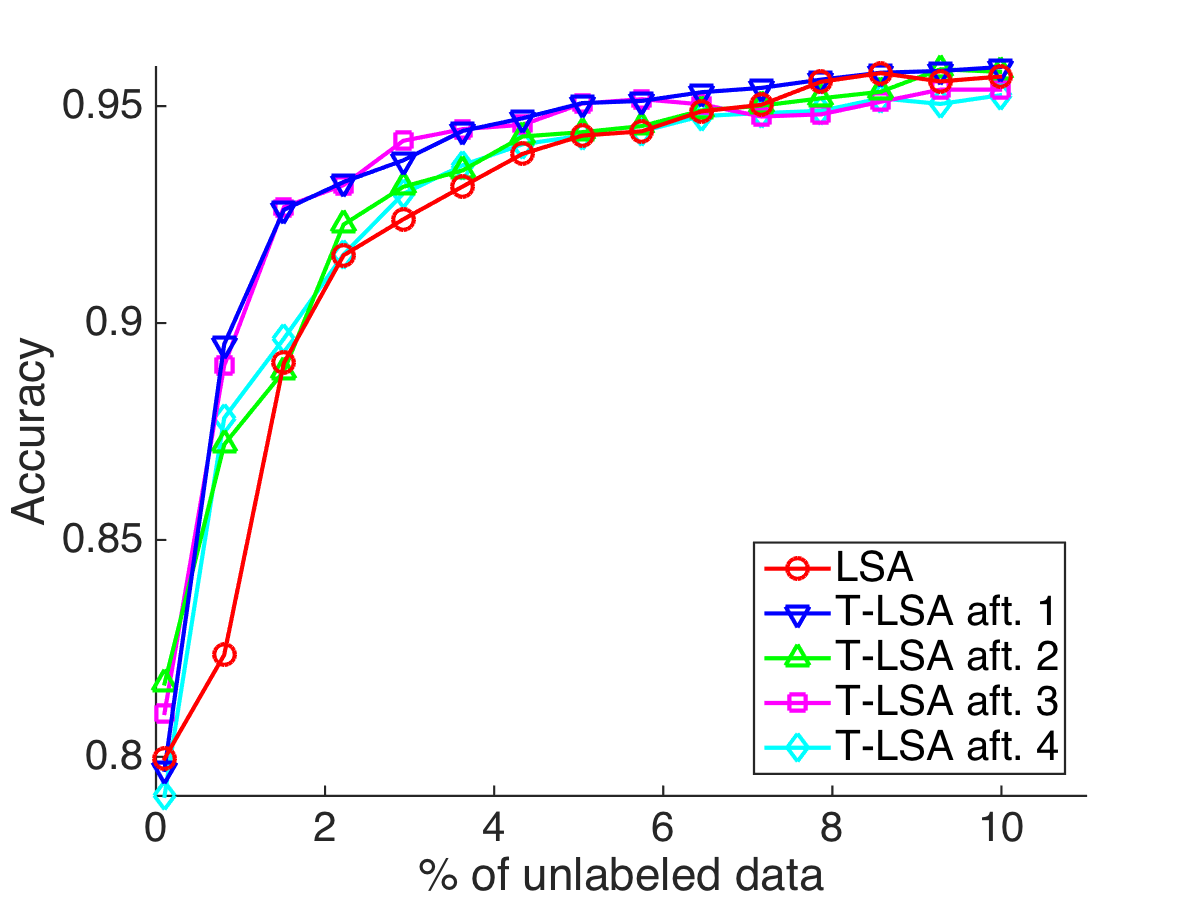}
                \caption{\textit{MNIST 8vs9}}
                \label{fig:MNIST_8vs9_lsa_tlsa_MNIST}
        \end{subfigure}%
\end{center}
\caption{Test Accuracy of LSA versus Transfer LSA on \textit{MNIST} and \textit{USPS}}
\label{fig:lsa_tlsa_homo}
\end{figure*}

\begin{figure*}[t]
\begin{center}
\centering
    \begin{subfigure}[b]{0.25\textwidth}
                \includegraphics[width=\textwidth]{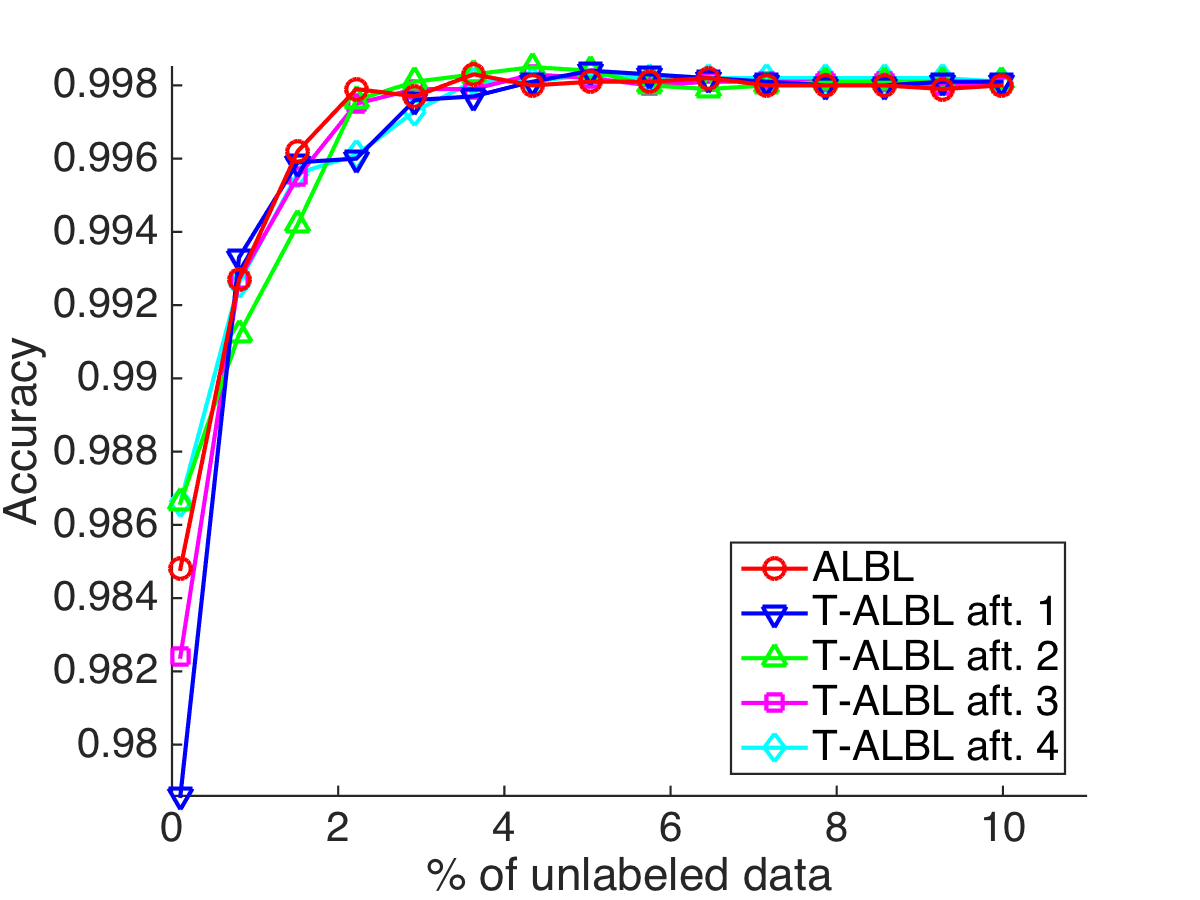}
                \caption{\textit{USPS 0vs1}}
                \label{fig:USPS_albl_talbl_USPS}
        \end{subfigure}%
    \begin{subfigure}[b]{0.25\textwidth}
                \includegraphics[width=\textwidth]{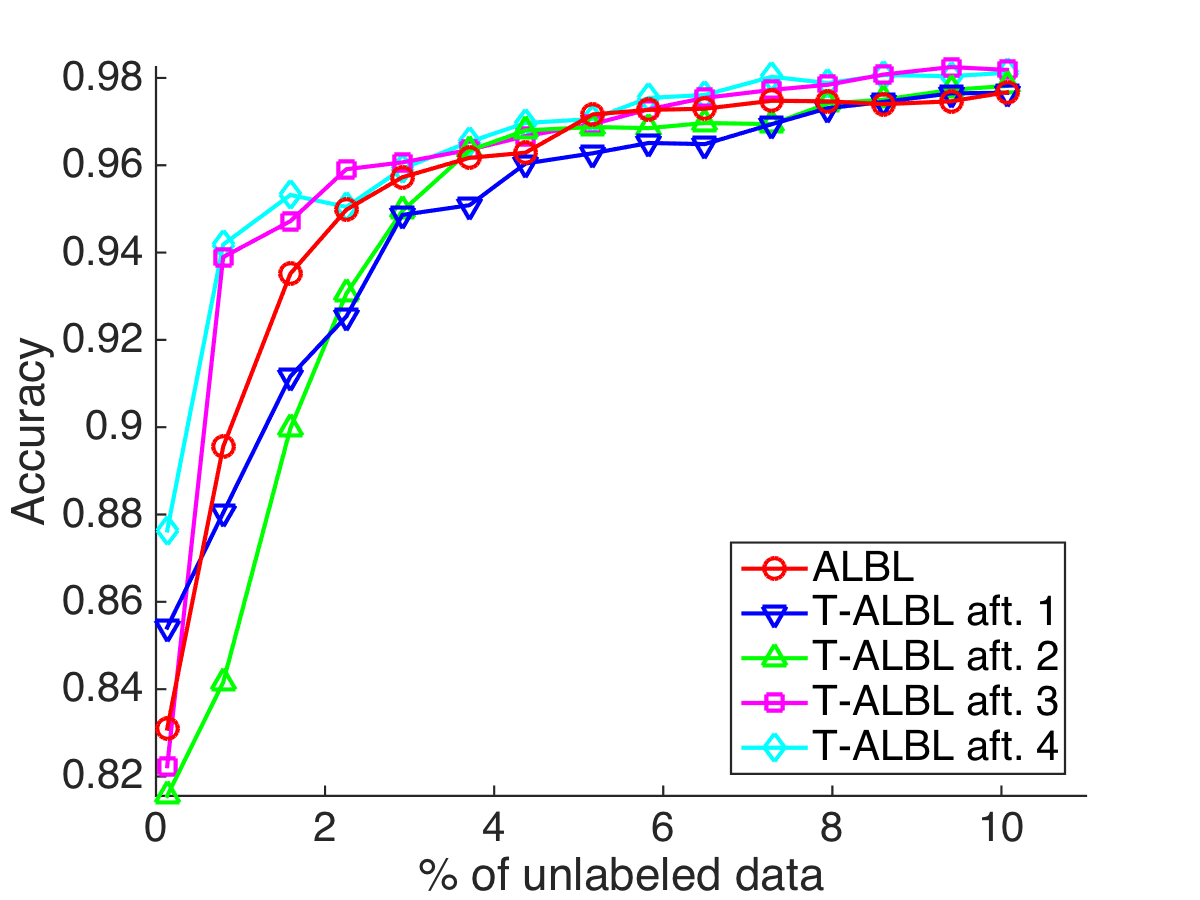}
                \caption{\textit{USPS 8vs9}}
                \label{fig:USPS_albl_talbl_USPS}
        \end{subfigure}%
    \begin{subfigure}[b]{0.25\textwidth}
                \includegraphics[width=\textwidth]{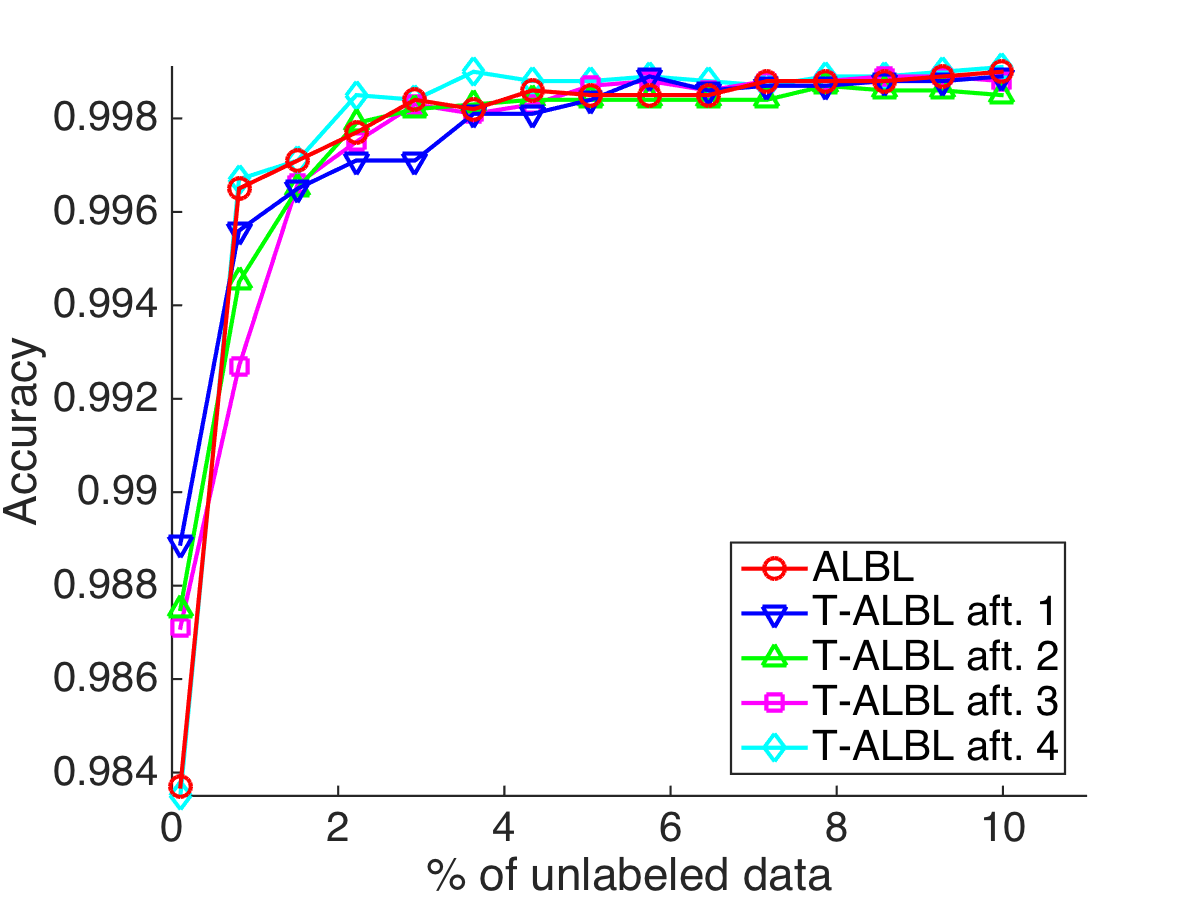}
                \caption{\textit{MNIST 0vs1}}
                \label{fig:MNIST_albl_talbl_MNIST}
        \end{subfigure}%
    \begin{subfigure}[b]{0.25\textwidth}
                \includegraphics[width=\textwidth]{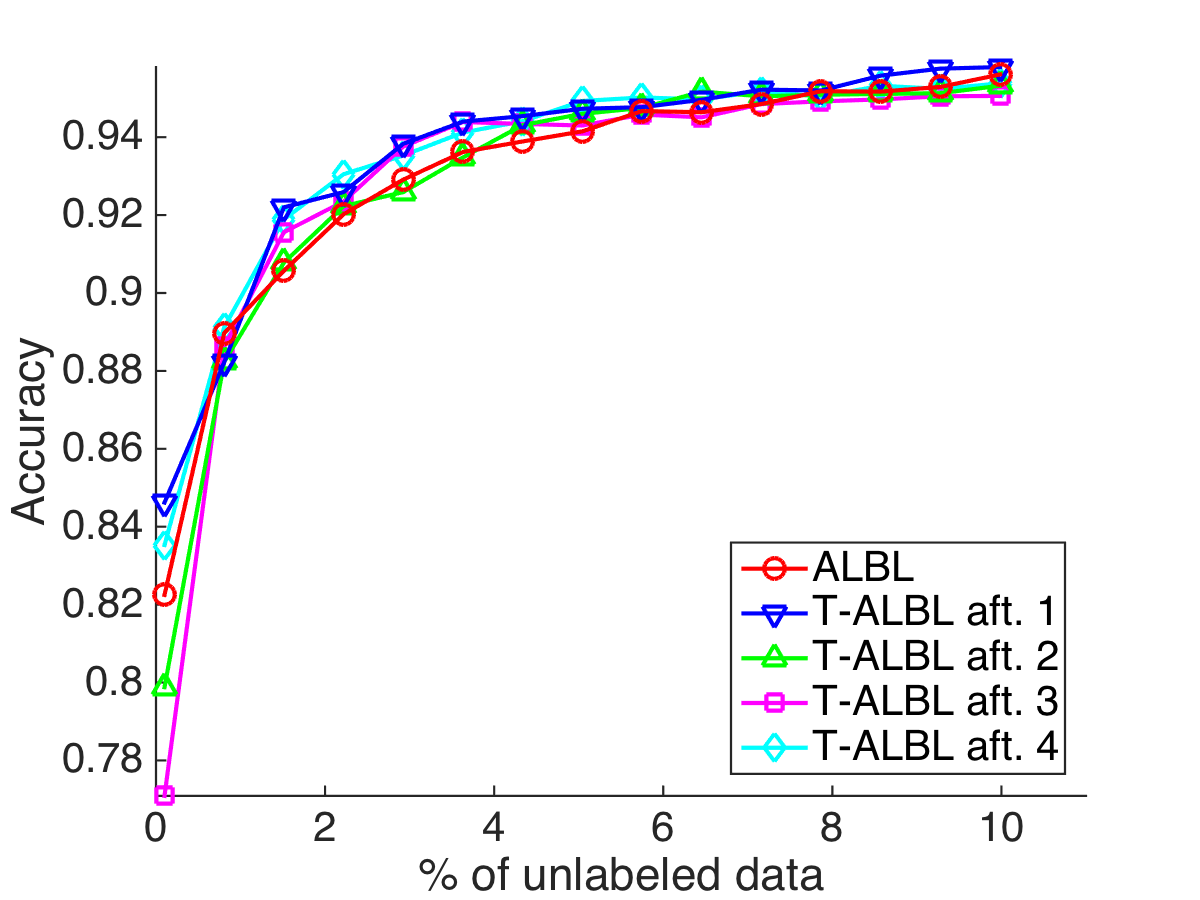}
                \caption{\textit{MNIST 8vs9}}
                \label{fig:MNIST_albl_talbl_MNIST}
        \end{subfigure}%

\end{center}
\caption{Test Accuracy of ALBL versus Transfer ALBL on \textit{USPS} and \textit{MNIST}}
\label{fig:albl_talbl_homo}
\end{figure*}

\begin{figure*}[t]
\begin{center}
\centering
    \begin{subfigure}[b]{0.25\textwidth}
                \includegraphics[width=\textwidth]{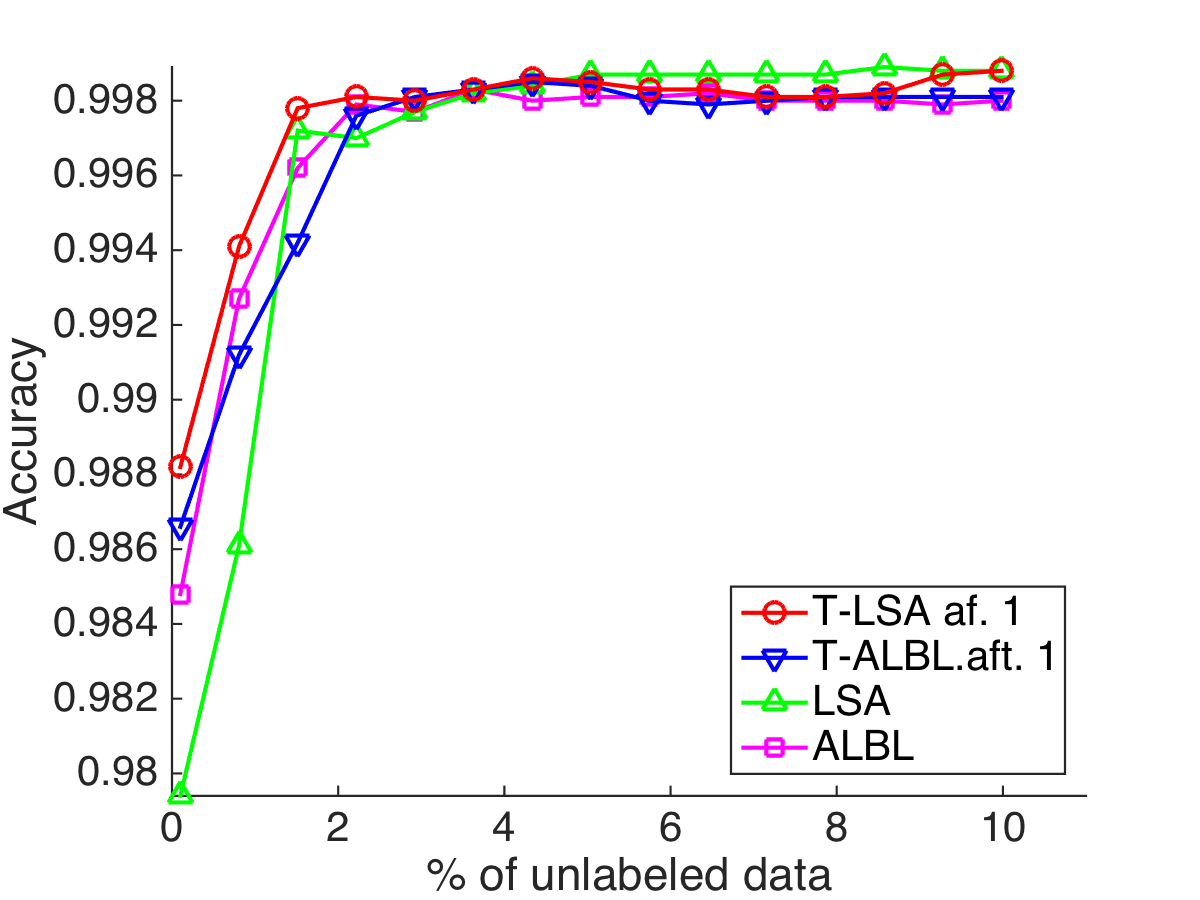}
                \caption{\textit{USPS 0vs1}}
                \label{fig:USPS_0vs1_tlsa_talbl_USPS}
        \end{subfigure}%
    \begin{subfigure}[b]{0.25\textwidth}
                \includegraphics[width=\textwidth]{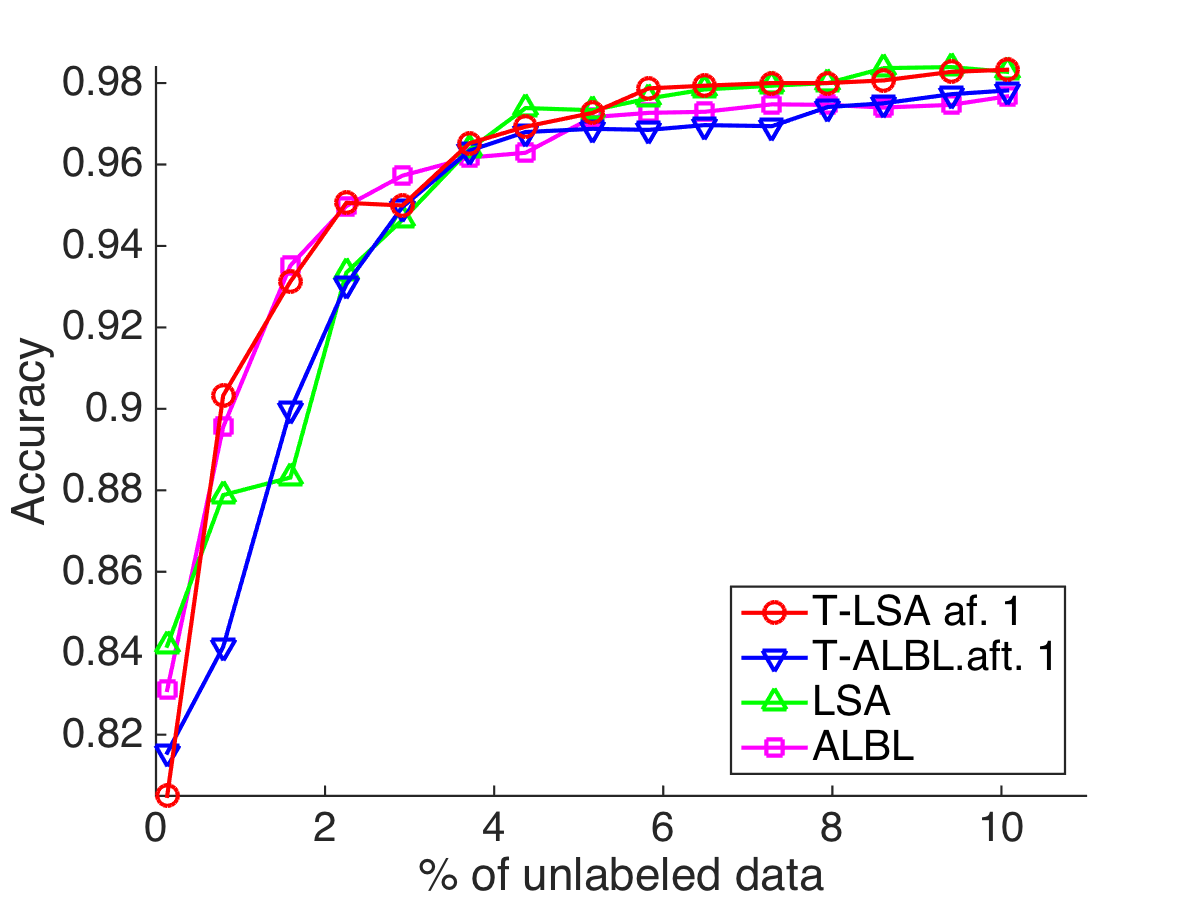}
                \caption{\textit{USPS 8vs9}}
                \label{fig:USPS_8vs9_tlsa_talbl_USPS}
        \end{subfigure}%
    \begin{subfigure}[b]{0.25\textwidth}
                \includegraphics[width=\textwidth]{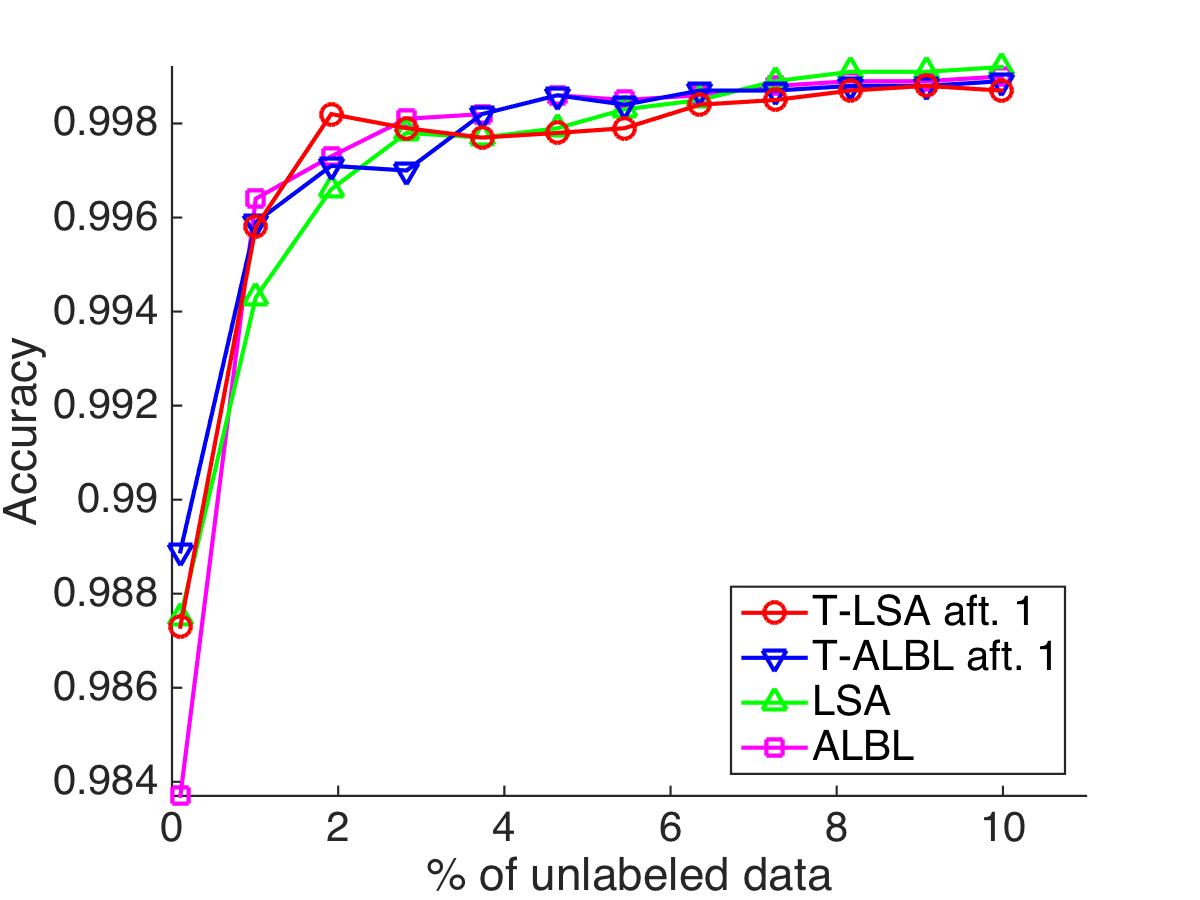}
                \caption{\textit{MNIST 0vs1}}
                \label{fig:MNIST_0vs1_tlsa_talbl_MNIST}
        \end{subfigure}%
    \begin{subfigure}[b]{0.25\textwidth}
                \includegraphics[width=\textwidth]{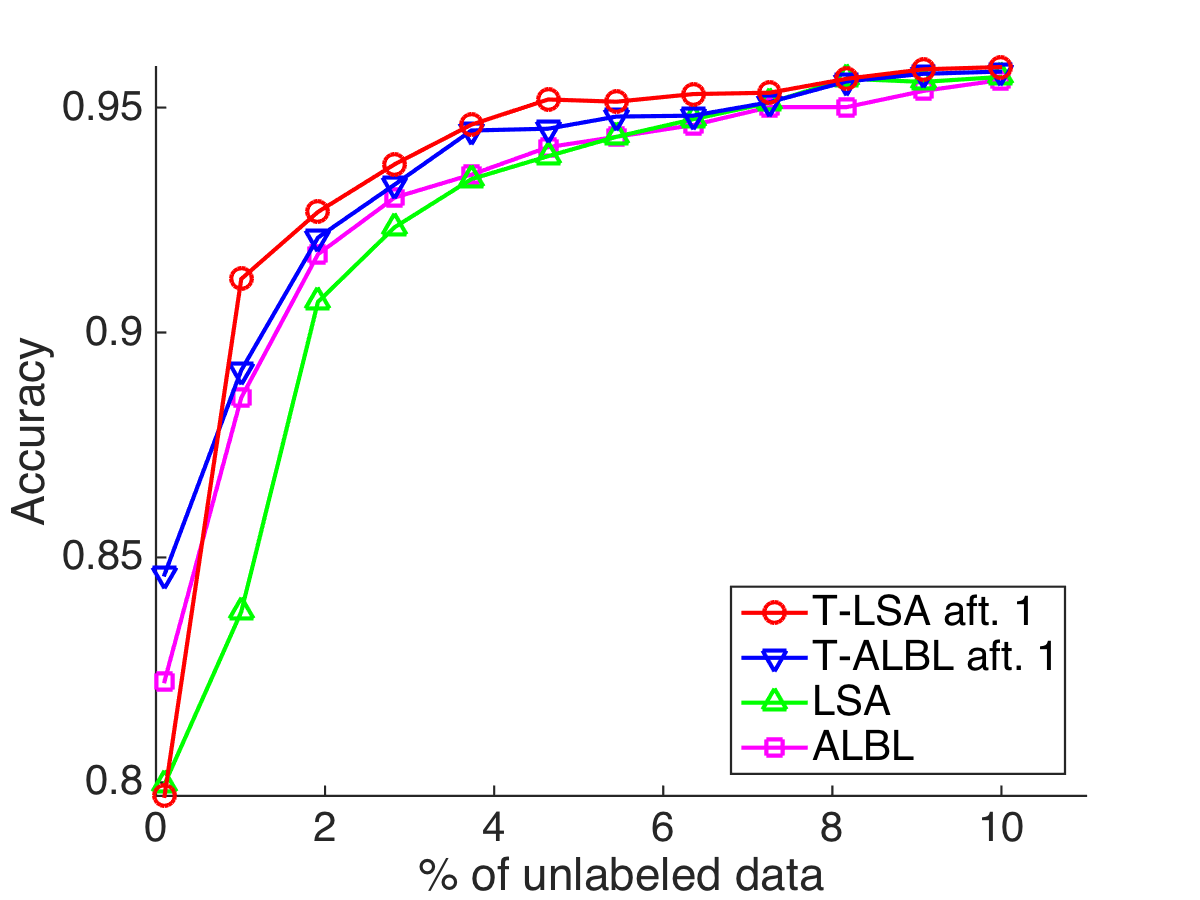}
                \caption{\textit{MNIST 8vs9}}
                \label{fig:MNIST_8vs9_tlsa_talbl_MNIST}
        \end{subfigure}%
\end{center}
\caption{Test Accuracy of Transfer LSA versus other competitors on \textit{USPS} and \textit{MNIST}}
\label{fig:tlsa_talbl_homo}
\end{figure*}

To understand the effectiveness of LSA as a blending approach, we compare LSA with ALBL. Because of space limits, we plot the test accuracy
along with the standard deviation on only four of the datasets, 
\textit{austra}, \textit{breast}, \textit{heart} and \textit{wdbc} in Fig. \ref{fig:lsa_albl}. 
We also compare LSA with ALBL with $t$-tests at $90\%$ confidence level on all datasets, and summarize the results in Table~\ref{tbl:lsa_albl}. 
The results of both Fig. \ref{fig:lsa_albl} and Table~\ref{tbl:lsa_albl} indicate that LSA is competitive to and sometimes even slightly better then ALBL. 
Furthermore, according to Fig.~\ref{fig:lsa_albl}, we can observe that 
the variation (standard deviation) of the LSA curve not only decreases more rapidly than that of the ALBL curve, but is also generally smaller after the first $10\%$ of the exploration queries.
The observation indicates that ALBL, being a probabilistic blending approach, is generally less stable than LSA, and matches our conjecture in Section~\ref{background} that the distribution in ALBL may be too volatile to
serve as robust active learning experience in practice.



\subsection{Experiments on Active Learning Across Datasets}\label{sec:exp_cross_datasets}

Next, we move to the experiments of cross-dataset active learning.
We first introduce the experiment setting before we proceed to discuss the details of the experiment results.
The experiment setting is as follows: 
A target dataset is first picked, and a random sequence that consists of other datasets is generated. 
Transferring algorithms, including T-LSA and T-ALBL, are then run on first $q$ datasets of the sequence to accumulate experience. 
With the previous experience, the active learning performance of the transferring algorithms is evaluated on the target dataset.
Each result is averaged over $10$ different random sequences.


The experiments of active learning across datasets are conducted in two different scenarios, 
where homogeneous and heterogeneous tasks are considered respectively. 
Specifically, a set of homogeneous tasks consists of datasets that share similar learning targets and the same feature space, 
and is constructed from the two benchmark datasets of multi-task learning.  
A set of heterogeneous tasks, on the other hand, involves datasets having different learning targets and feature space, 
and is simulated by the eight benchmark datasets of active learning. 
We will first discuss the experiments on homogeneous tasks, 
where algorithms that exploit the transferred experience are expected to perform better.
The experiments on heterogeneous tasks, which is a more general but more challenging scenario, will then be discussed.

For the experiments in each scenario, we first compare T-LSA and T-ALBL using experience from different number of previous datasets (i.e. different $q$) with their non-transferring predecessors, 
namely LSA and ALBL, 
to evaluate the effectiveness of experience transfer of active learning.  
Then, we will directly compare T-LSA with T-ALBL, LSA and ALBL using a specific $q$ to understand the absolute performance difference 
between T-LSA and other competitors. 

We choose to not include QUIRE in the cross-dataset experiments because QUIRE is considerably more time-consuming given its label-assignment estimation steps.

\paragraph{Experiments on Homogeneous Tasks}\label{sec:exp_ho_datasets}

The experiments of learning across homogeneous tasks are conducted on two benchmark datasets of hand-written digit recognition, \textit{USPS} and \textit{MNIST}, for multi-task learning. 
We split both \textit{USPS} and \textit{MNIST} into 5 binary classification datasets, 
namely \textit{0vs1}, \textit{2vs3}, \textit{4vs5}, \textit{6vs7} and \textit{8vs9} to construct the set of homogeneous learning tasks. 
Since the active learning performance on \textit{USPS} and \textit{MNIST} converges quickly, 
we only compare the results with respect to the queries in first $10\%$ of unlabeled data to better illustrate the difference.

We compare T-LSA with LSA and T-ALBL with ALBL, and present the results of test accuracy in 
Fig.~\ref{fig:lsa_tlsa_homo} and Fig.~\ref{fig:albl_talbl_homo} respectively. 
Owing to the readability, only results of two tasks on each dataset are presented here.
From Fig.~\ref{fig:lsa_tlsa_homo}, we observe that T-LSA generally outperforms LSA on tasks \textit{0vs1} and \textit{8vs9} of both \textit{USPS} and \textit{MNIST}.
On the other hand, Fig.~\ref{fig:albl_talbl_homo} indicates that T-ALBL performs similarly or even worse than ALBL on task \textit{0vs1} of both \textit{USPS} and \textit{MNIST}. 
For task \textit{8vs9}, the improvement of T-ALBL with regard to ALBL is rather minor on \textit{MNIST}, while obvious negative transfer can be observed on \textit{USPS}. 

We then compare T-LSA with T-ALBL, LSA, and ALBL directly using experience from one previous dataset,
to examine the absolute performance difference between T-LSA and other competitors. 
The results are illustrated in Fig.~\ref{fig:tlsa_talbl_homo}. 
These competitors are further compared on all five tasks of both datasets based on $t$-test at $90\%$ confidence level, and the results are summarized in Table~\ref{tbl:tlsa_talbl_homo}. 
From Fig. \ref{fig:tlsa_talbl_homo}, we can observe that performance of LSA is again inferior especially in the first $4\%$ of queries. 
T-LSA, on the other hand, often performs the best among all four competitors.
The results of Table \ref{tbl:tlsa_talbl_homo} indicates a slight improvement of T-LSA over LSA in the initial stage of learning, and shows the competitive performance of T-LSA over other competitors.

The observations on both \textit{USPS} and \textit{MNIST} demonstrate that T-LSA successfully improves the active learning performance of LSA by 
transferring the active learning experience via the proposed linear weights,
which is as expected in the scenario of active learning across homogeneous tasks.
\mbox{T-ALBL}, however, often performs inferior than ALBL,
confirming that experience transfer via the probability distribution of ALBL can lead to negative impact.

\begin{table}[t]
\caption{Transfer LSA versus other competitors on USPS and MNIST based on $t$-test at 90\% confidence level (\#win/\#tie/\#loss)}
\label{tbl:tlsa_talbl_homo}
\begin{center}
\resizebox{0.48\textwidth}{!}{%
\begin{tabular}{l c c c c c l}
\hline 
& \multicolumn{5}{c}{\bf percentage of queried instances} & \multirow{2}{*}{total}\\
\cline{2-6}
& 2\% & 4\% & 6\% & 8\% & 10\% &\\
\hline \hline 
LSA&             2/8/0 & 0/10/0 & 0/10/0 & 0/10/0 & 0/10/0 & 2/48/0 \\
ALBL&            0/9/1 & 0/10/0 & 2/8/0 & 0/10/0 & 1/9/0 & 3/46/1 \\
T-ALBL aft. 1&   0/10/0 & 0/9/1 & 1/8/1 & 1/8/1 & 1/9/0 & 3/44/3 \\
\hline
 total&          2/27/1 & 0/29/1 & 3/26/1 & 1/28/1 & 2/28/0 & 8/138/4 \\
\hline 
\end{tabular}
}
\end{center}
\end{table}

\begin{figure*}[t]
\begin{center}
\centering
    \begin{subfigure}[b]{0.25\textwidth}
                \includegraphics[width=\textwidth]{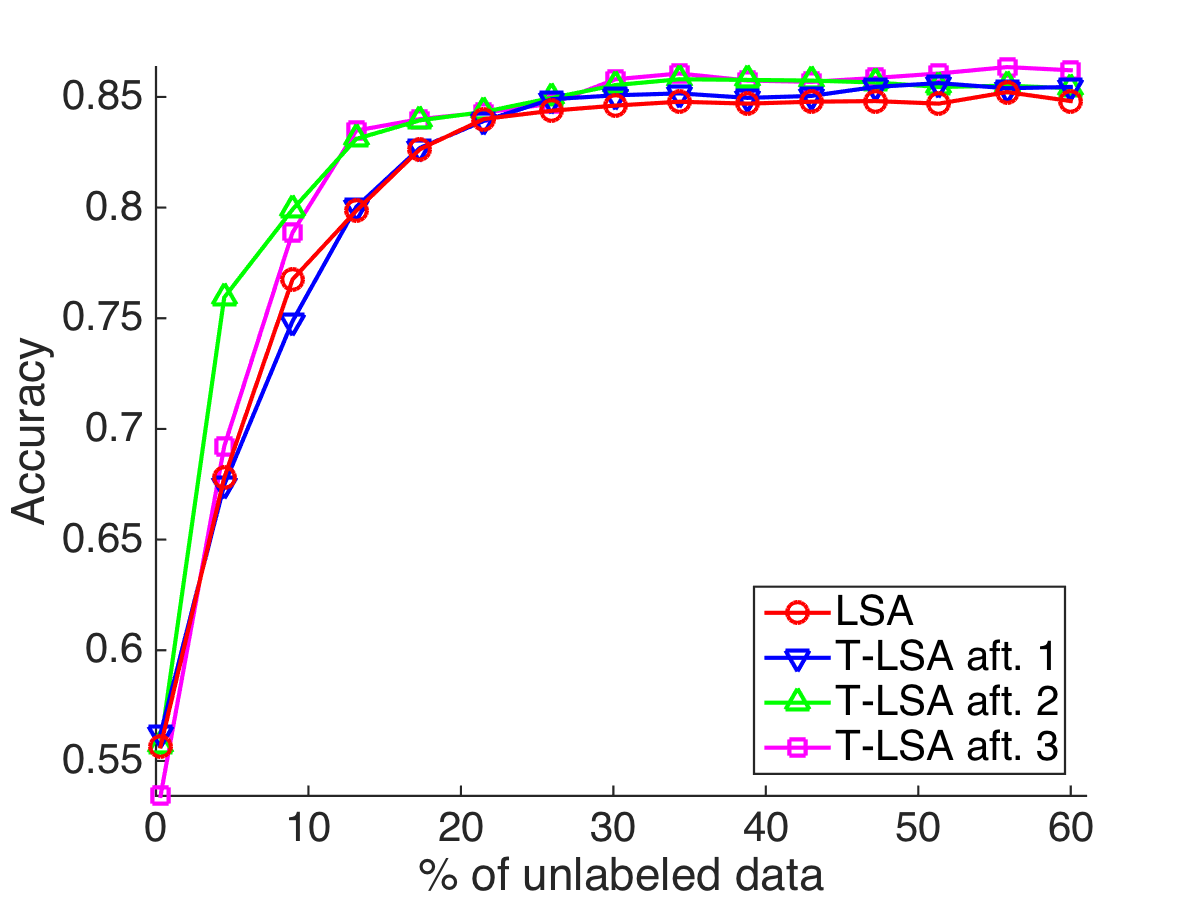}
                \caption{\textit{austra}}
                \label{fig:austra_lsa_tlsa}
        \end{subfigure}%
    \begin{subfigure}[b]{0.25\textwidth}
                \includegraphics[width=\textwidth]{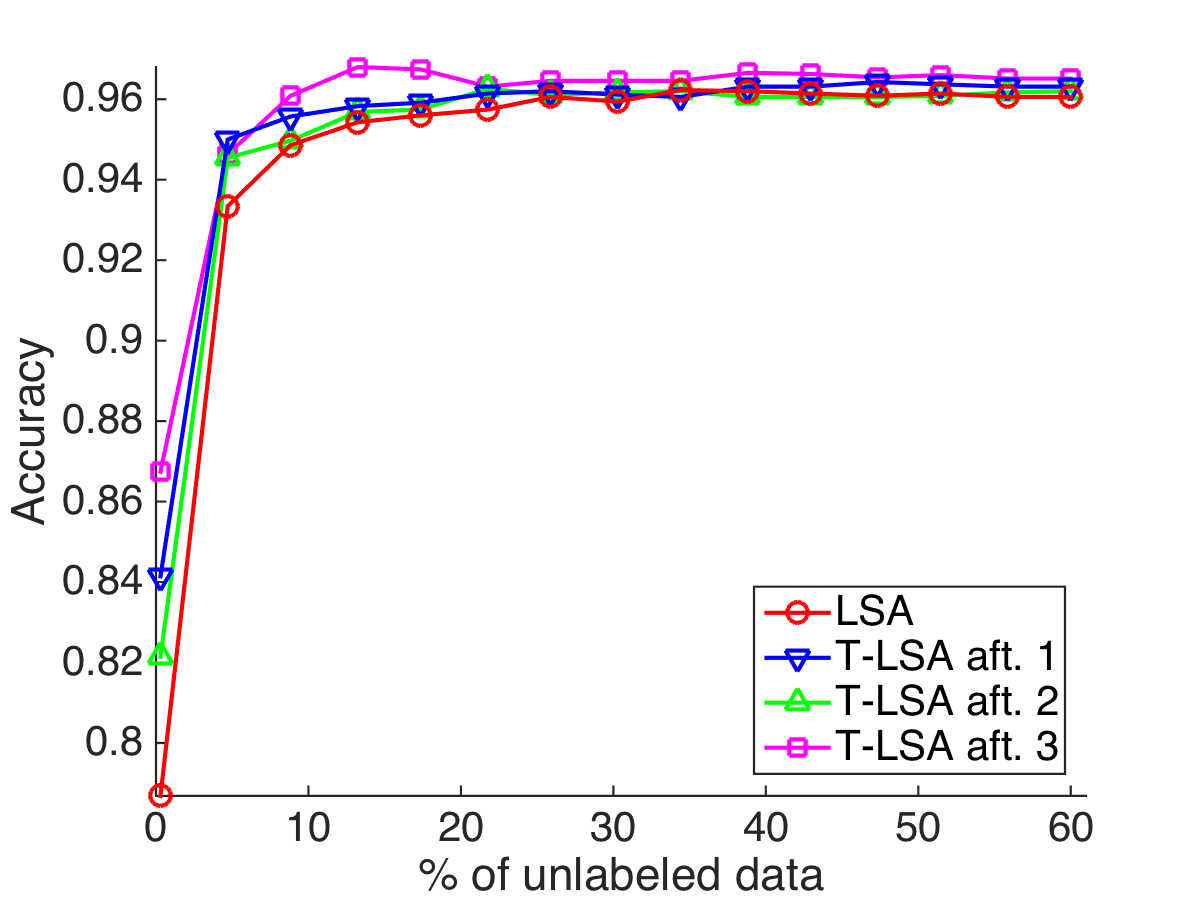}
                \caption{\textit{breast}}
                \label{fig:breast_lsa_tlsa}
        \end{subfigure}%
    \begin{subfigure}[b]{0.25\textwidth}
                \includegraphics[width=\textwidth]{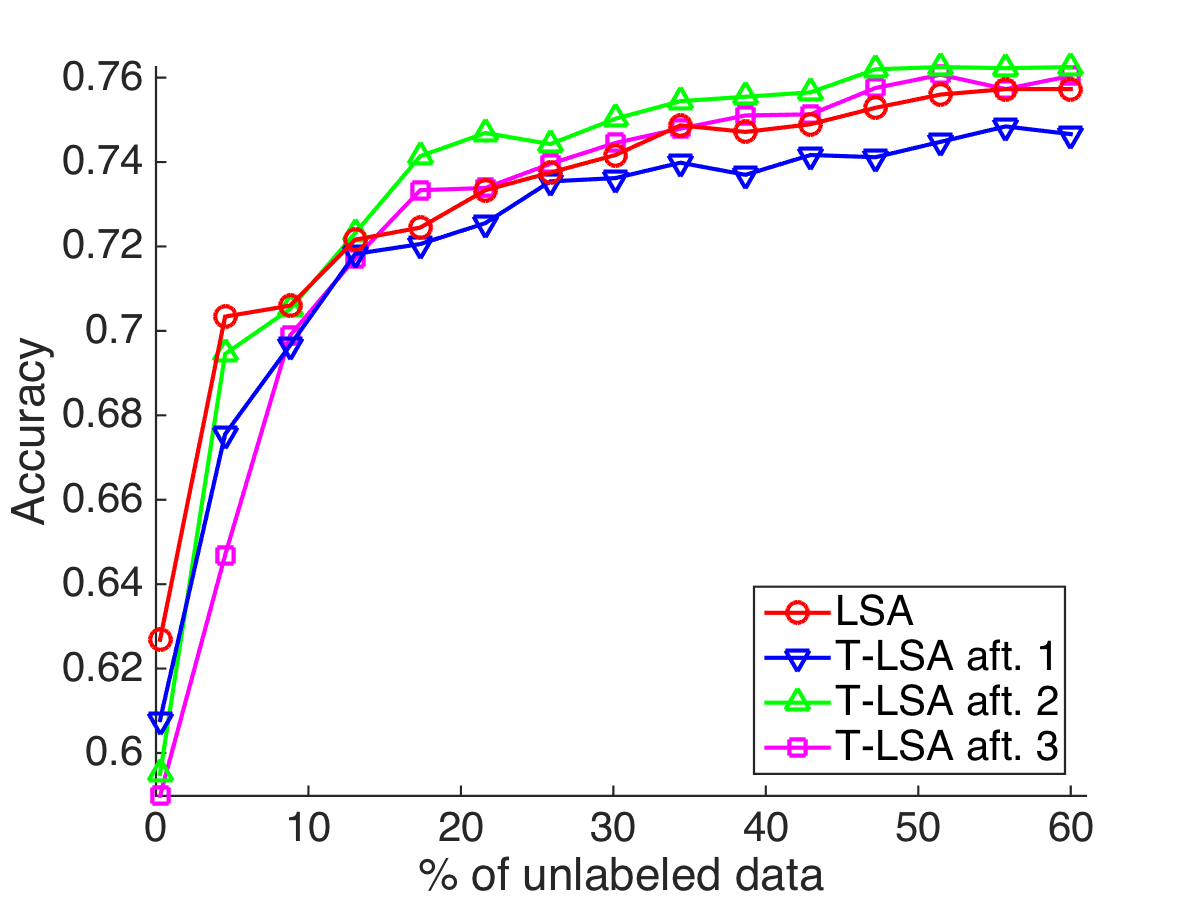}
                \caption{\textit{diabetes}}
                \label{fig:diabetes_lsa_tlsa}
        \end{subfigure}%
    \begin{subfigure}[b]{0.25\textwidth}
                \includegraphics[width=\textwidth]{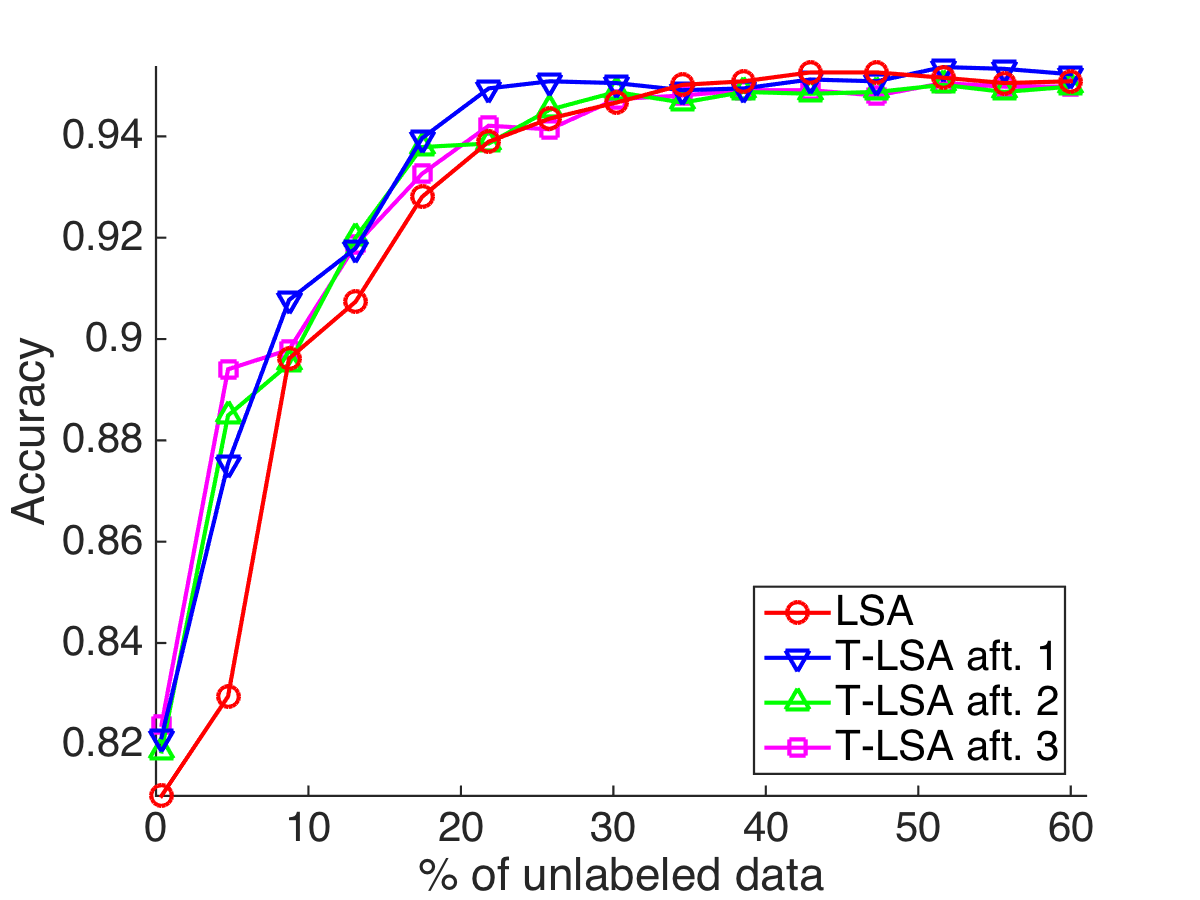}
                \caption{\textit{wdbc}}
                \label{fig:wdbc_lsa_tlsa}
        \end{subfigure}%

\end{center}
\caption{Test Accuracy of LSA versus Transfer LSA}
\label{fig:lsa_tlsa}
\end{figure*}

\begin{figure*}[t]
\begin{center}
\centering
    \begin{subfigure}[b]{0.25\textwidth}
                \includegraphics[width=\textwidth]{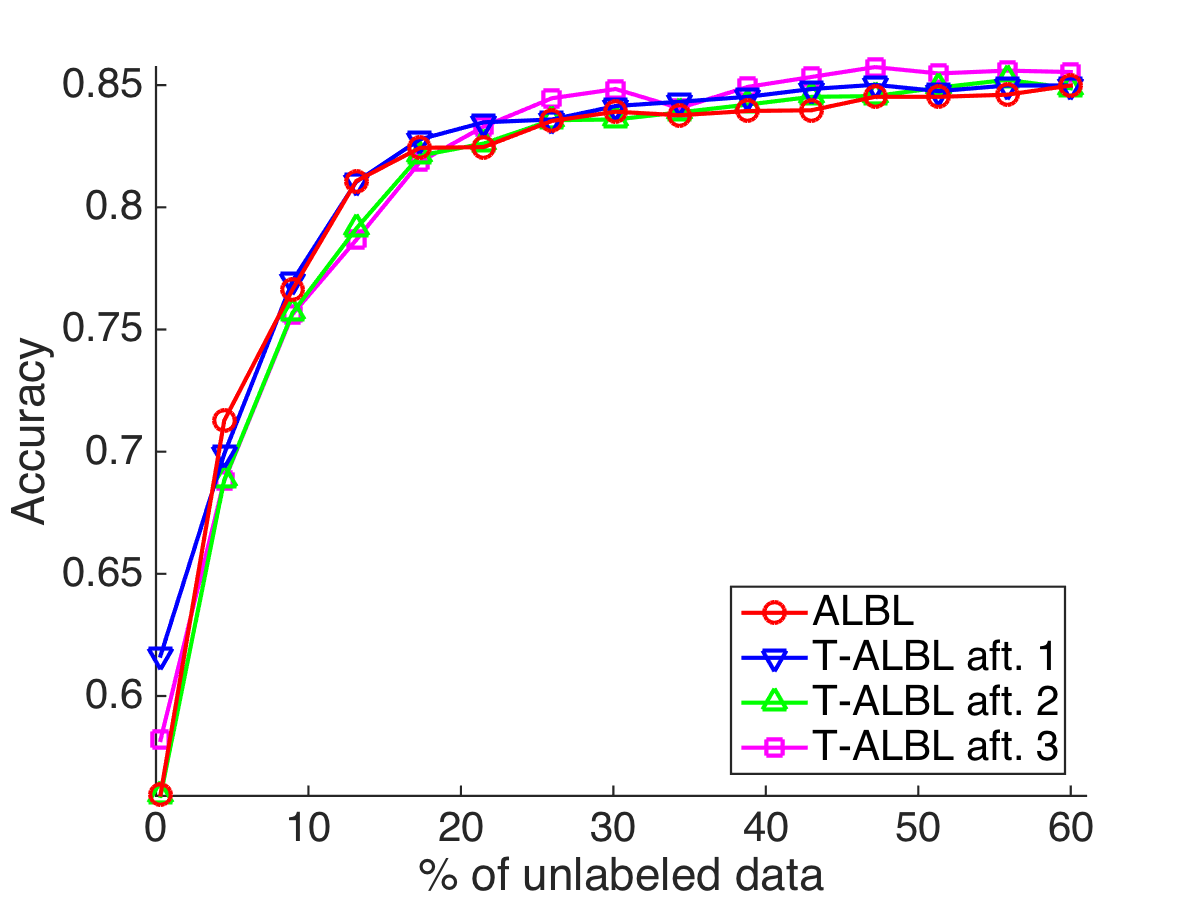}
                \caption{\textit{austra}}
                \label{fig:austra_albl_talbl}
        \end{subfigure}%
    \begin{subfigure}[b]{0.25\textwidth}
                \includegraphics[width=\textwidth]{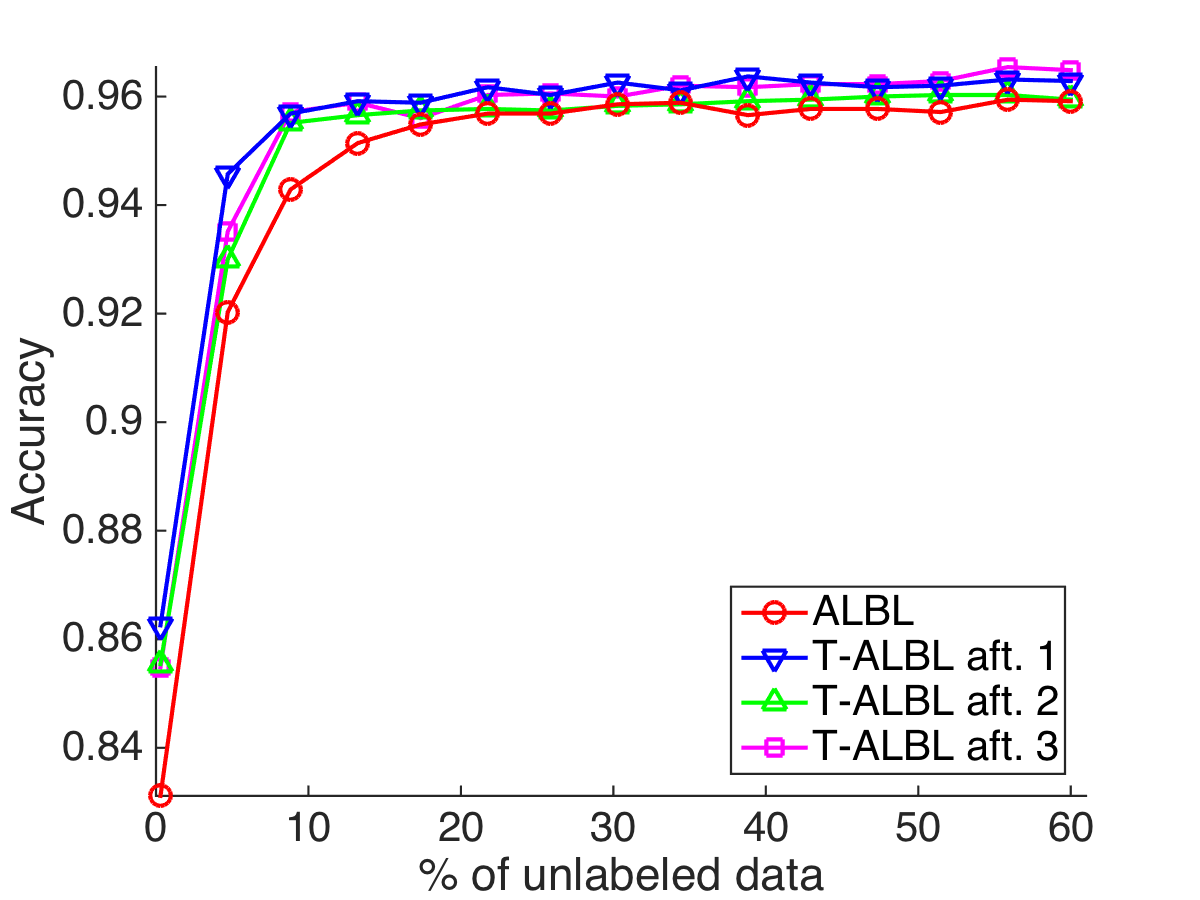}
                \caption{\textit{breast}}
                \label{fig:breast_albl_talbl}
        \end{subfigure}%
    \begin{subfigure}[b]{0.25\textwidth}
                \includegraphics[width=\textwidth]{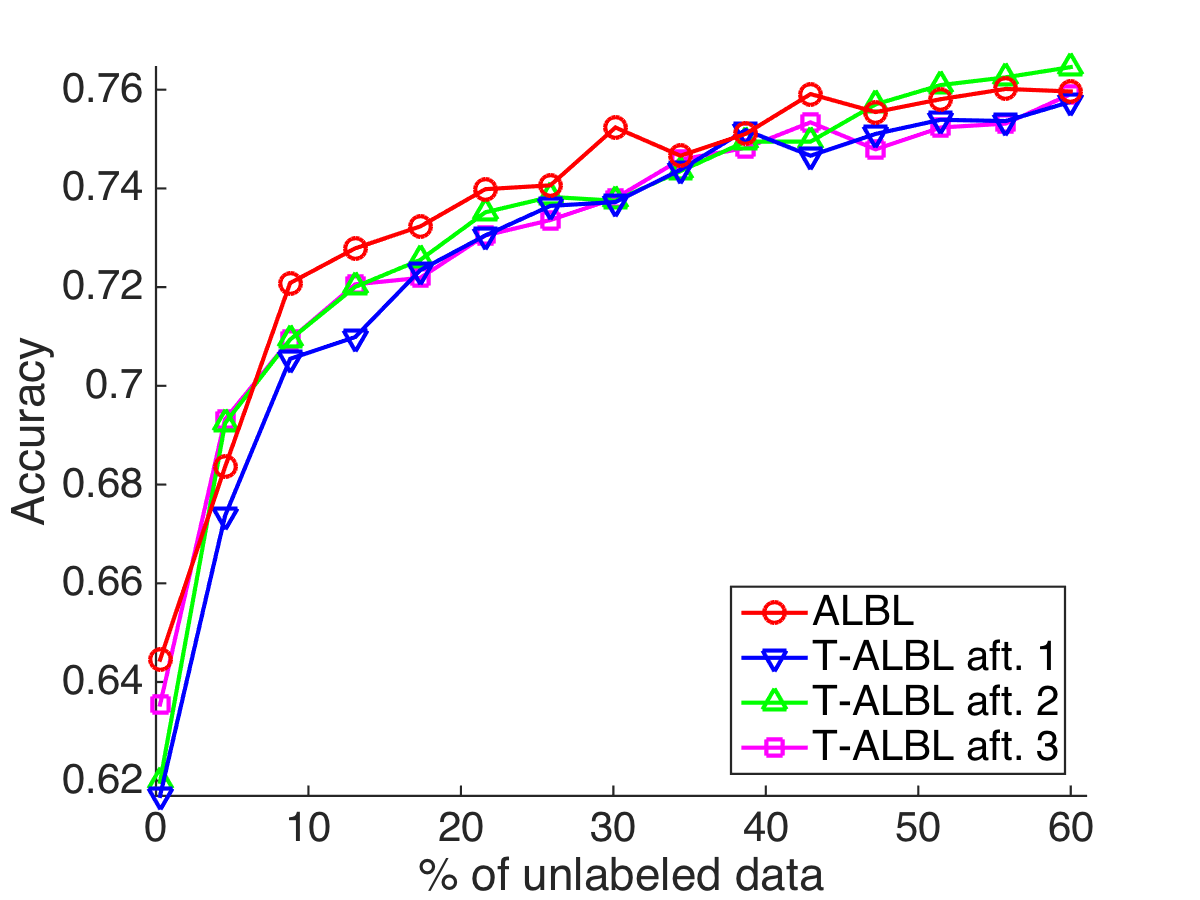}
                \caption{\textit{diabetes}}
                \label{fig:diabetes_albl_talbl}
        \end{subfigure}%
    \begin{subfigure}[b]{0.25\textwidth}
                \includegraphics[width=\textwidth]{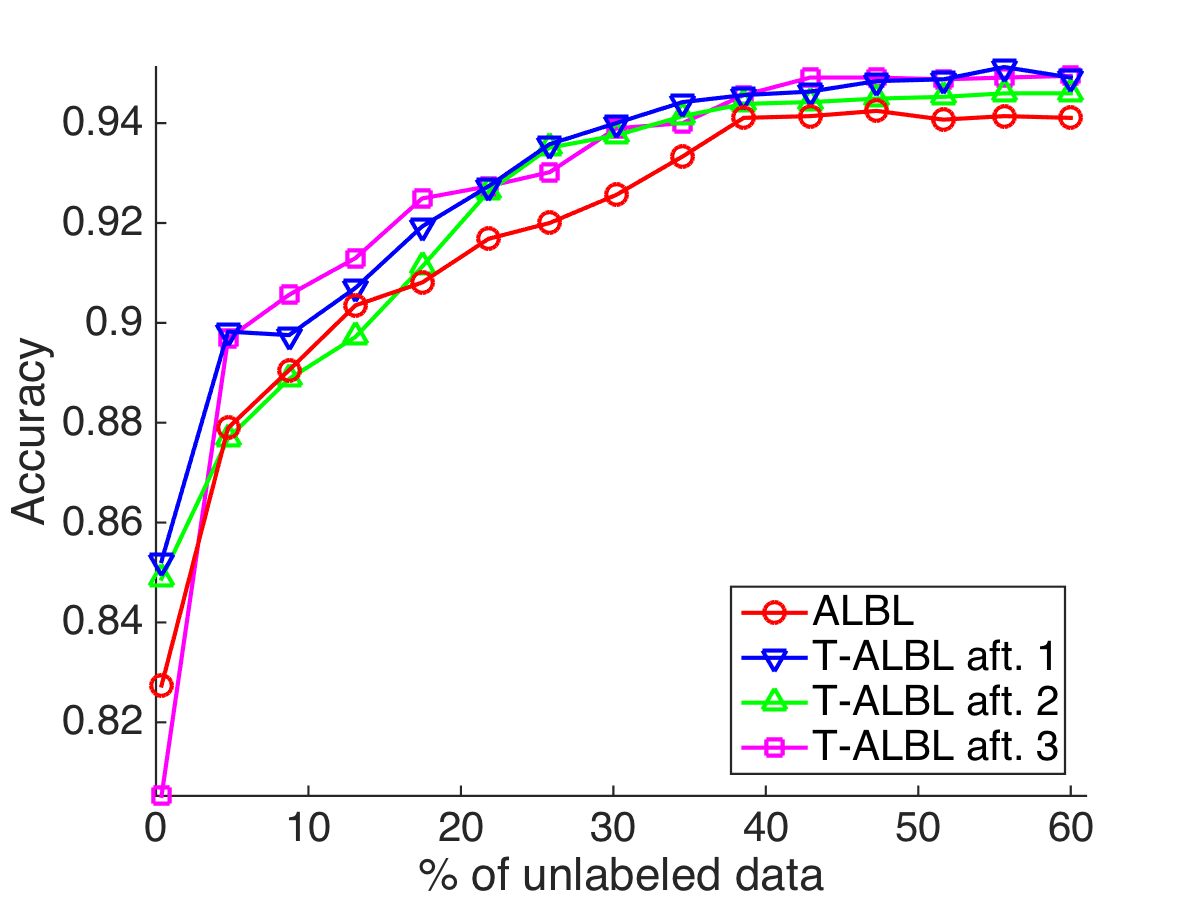}
                \caption{\textit{wdbc}}
                \label{fig:wdbc_albl_talbl}
        \end{subfigure}%

\end{center}
\caption{Test Accuracy of ALBL versus Transfer ALBL}
\label{fig:albl_talbl}
\end{figure*}

\begin{figure*}[t]
\begin{center}
\centering
    \begin{subfigure}[b]{0.25\textwidth}
                \includegraphics[width=\textwidth]{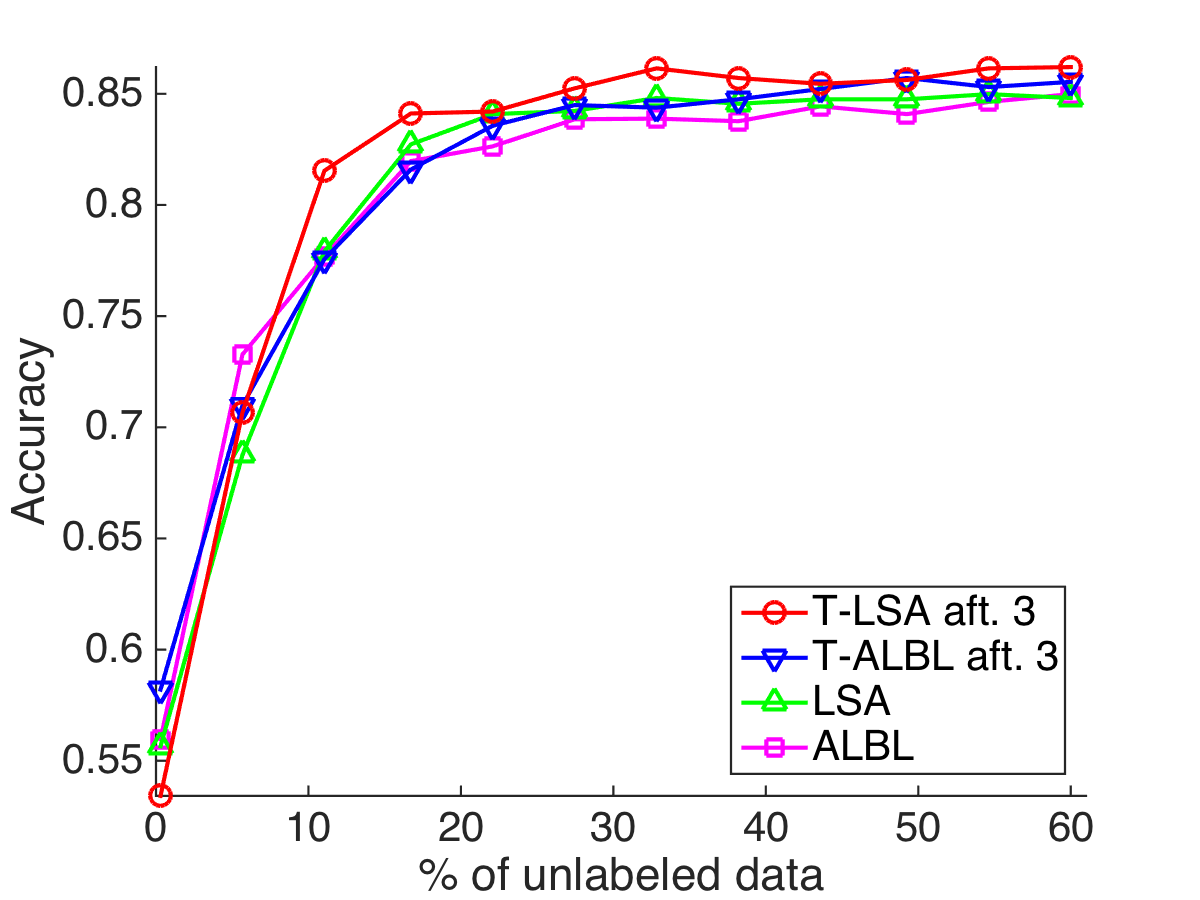}
                \caption{\textit{austra}}
                \label{fig:austra_tlsa_talbl}
        \end{subfigure}%
    \begin{subfigure}[b]{0.25\textwidth}
                \includegraphics[width=\textwidth]{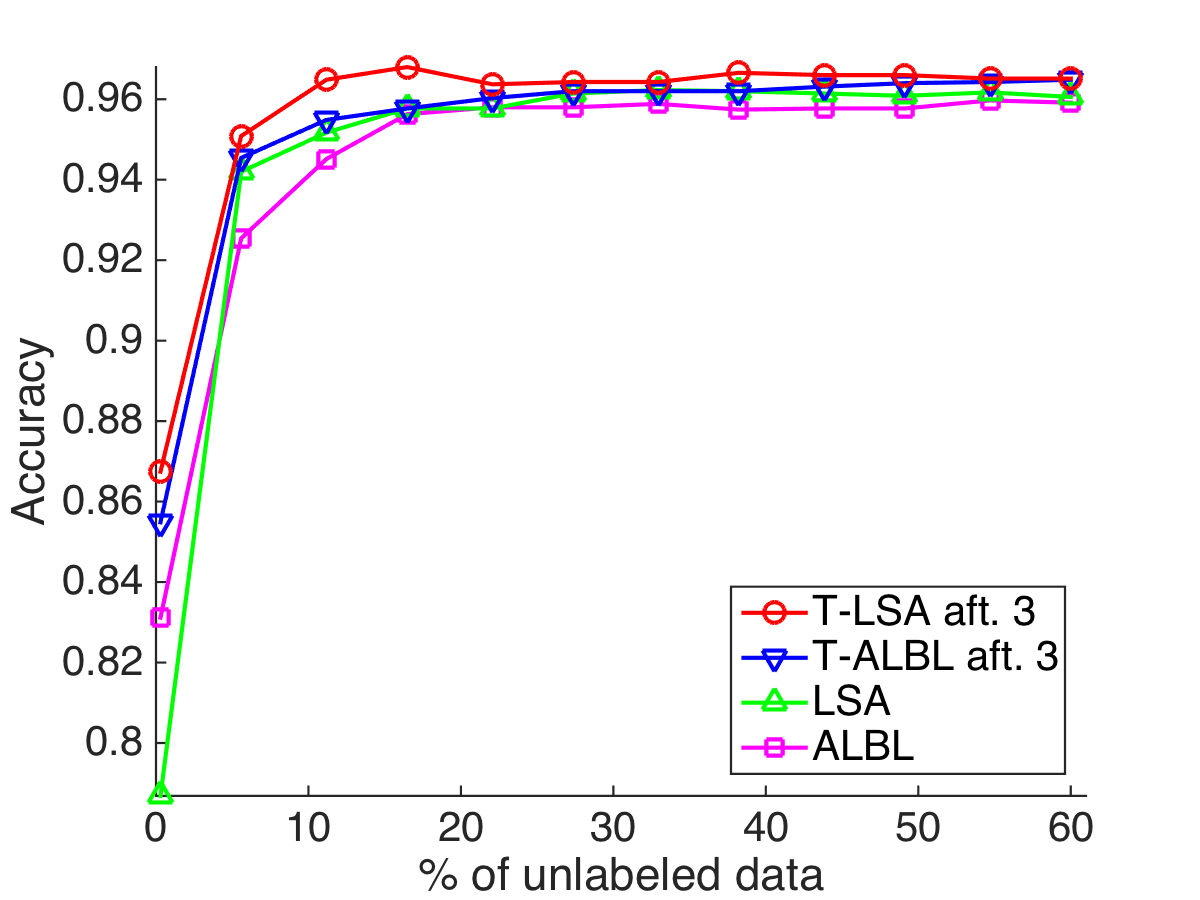}
                \caption{\textit{breast}}
                \label{fig:breast_tlsa_talbl}
        \end{subfigure}%
    \begin{subfigure}[b]{0.25\textwidth}
                \includegraphics[width=\textwidth]{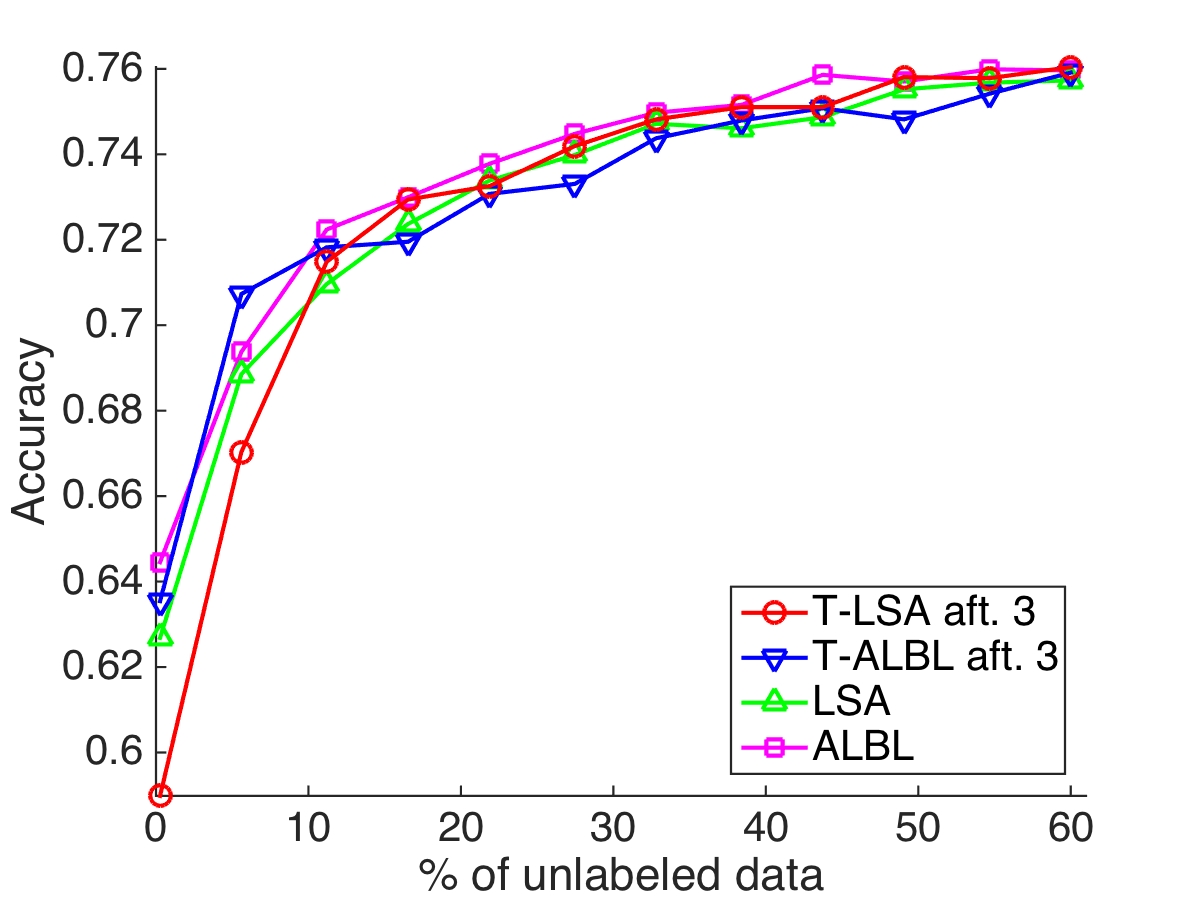}
                \caption{\textit{diabetes}}
                \label{fig:diabetes_tlsa_talbl}
        \end{subfigure}%
    \begin{subfigure}[b]{0.25\textwidth}
                \includegraphics[width=\textwidth]{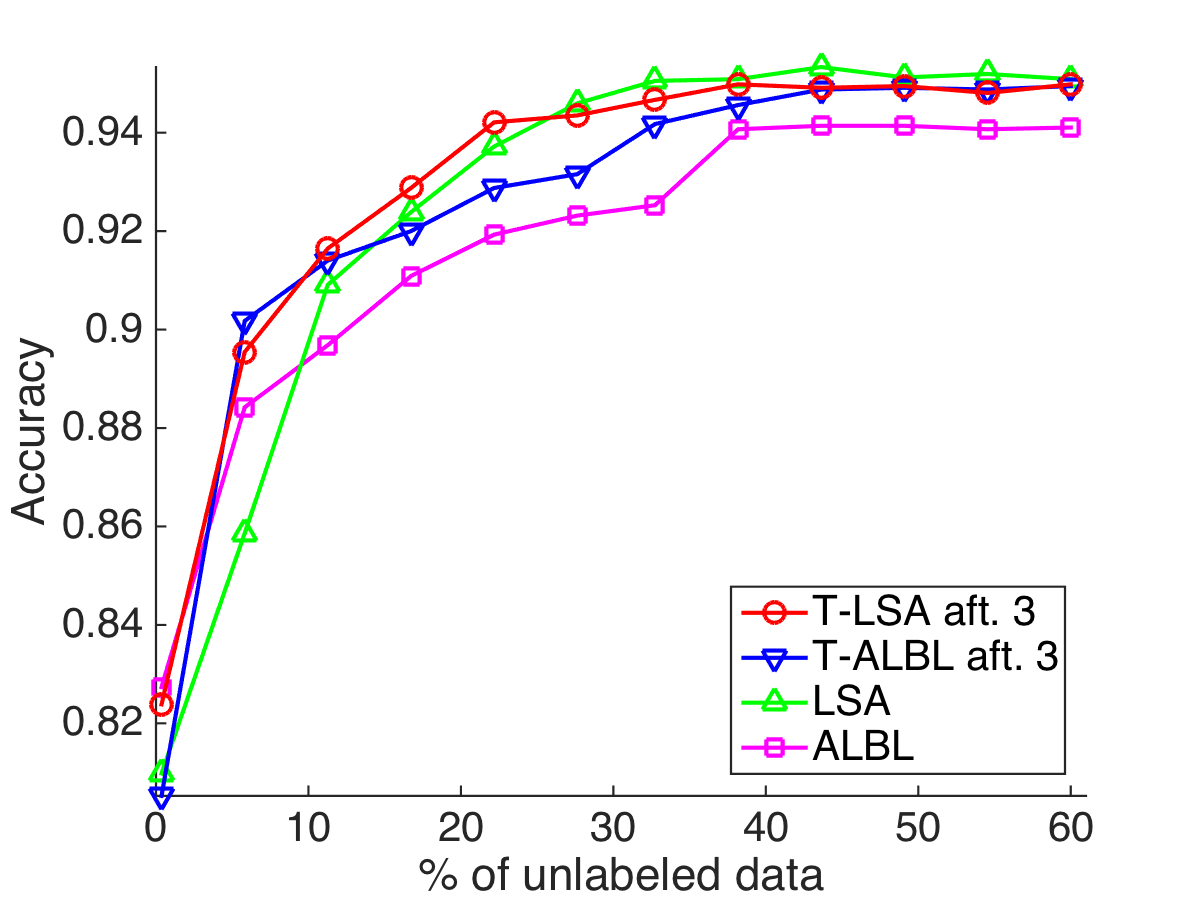}
                \caption{\textit{wdbc}}
                \label{fig:german_tlsa_talbl}
        \end{subfigure}%

\end{center}
\caption{Test Accuracy of Transfer LSA versus Transfer ALBL}
\label{fig:tlsa_talbl}
\end{figure*}

\paragraph{Experiments on Heterogeneous Tasks}\label{sec:exp_he_datasets}

Next, we shall discuss the experiments on learning across heterogeneous tasks.
The experiments are conducted on the eight benchmark datasets of active learning.
The feature spaces and the learning targets vary from each others between different active learning datasets.
We set $q = 3$ in the experiments. 

We first compare T-LSA with LSA and T-ALBL with ALBL, 
and present the results in Fig. \ref{fig:lsa_tlsa} and Fig. \ref{fig:albl_talbl} respectively.  
Owing to the space limits and readability, only selected results on \textit{austra}, \textit{breast}, \textit{diabetes} and \textit{wdbc} are presented.
According to Fig. \ref{fig:lsa_tlsa}, T-LSA improves the performance of LSA on datasets \textit{austra}, \textit{breast} and \textit{wdbc}. For dataset \textit{diabetes}, T-LSA is inferior in the initial stage, but 
can quickly catch up and even outperform LSA.
On the other hand, we can observe from Fig.~\ref{fig:albl_talbl} that T-ALBL improves over ALBL on datasets \textit{breast} and \textit{wdbc}, but is inferior on datasets \textit{austra} and \textit{diabetes}. 

\begin{table}[t]
\caption{Transfer LSA versus other competitors based on $t$-test at 90\% confidence level (\#win/\#tie/\#loss)}
\label{tbl:tlsa_talbl}
\begin{center}
\resizebox{0.48\textwidth}{!}{%
\begin{tabular}{l c c c c c c c l}
\hline 
 & \multicolumn{7}{c}{\bf percentage of queried instances} & \multirow{2}{*}{total}\\
\cline{2-8}
& 5\% & 10\% & 15\% & 20\% & 30\% & 40\% & 50\% &\\
\hline \hline 
LSA& 1/7/0 & 1/7/0 & 1/7/0 & 1/7/0 & 1/7/0 & 2/6/0 & 1/6/1 & 8/47/1 \\
ALBL& 0/7/1 & 2/6/0 & 2/6/0 & 3/5/0 & 3/5/0 & 3/5/0 & 2/6/0 & 15/40/1 \\
T-ALBL aft. 3& 0/7/1 & 0/8/0 & 2/6/0 & 3/5/0 & 1/7/0 & 1/7/0 & 1/7/0 & 8/47/1 \\
\hline
 total& 1/21/2 & 3/21/0 & 5/19/0 & 7/17/0 & 5/19/0 & 6/18/0 & 4/19/1 & 31/134/3 \\
\hline 
\end{tabular}
}
\end{center}
\end{table}

We then compare T-LSA directly with T-ALBL, LSA and ALBL, 
where the transferring algorithms can exploit the experience from previous $3$ datasets (i.e. $q$ = 3). We 
illustrate the results in Fig. \ref{fig:tlsa_talbl}.
We also compare these algorithms based on $t$-test at $90\%$ confidence level, and summarize the results in Table \ref{tbl:tlsa_talbl}. 
From Fig. \ref{fig:tlsa_talbl}, T-LSA reaches the best performance among all four competitors on datasets \textit{austra}, \textit{breast} and \textit{wdbc}, and can catch up with the best competitor after querying $10\%$ of unlabeled data on dataset \textit{diabetes}. 
The results of Table \ref{tbl:tlsa_talbl} further confirm that T-LSA can often outperform other competitors.

The aforementioned observations demonstrate that  
experience transfer via our proposed linear weights is superior to that via the probabilistic distribution of ALBL with the following two advantages: 
(1) better improvement from experience transfer and (2) ability to recover more quickly when the transferred experience is negative to performance.
In addition, T-LSA is shown to improve over LSA by providing a better starting point
for exploration in the initial stage of active learning.

The success of T-LSA in both scenarios positively answers the question in our title, 
where active learning experience can indeed be transferred to improve the active learning performance.



\section{Conclusion} \label{conclusion}

We propose a novel approach that accomplishes the mission of transferring active learning experience across datasets.
The approach is based on a unified representation of human knowledge and environment status about active learning,
and a linear model on the representation. The model allows taking the linear weights as experience, and can be updated
by the LinUCB algorithm for contextual bandit learning through a novel reward function. The experience learned from the model
can be transferred to other active learning tasks through biased regularization. Empirical studies not only confirm the competitiveness
of the proposed approach, but also confirm that it can be beneficial to transfer the experience across active learning tasks that are either homogeneous or heterogeneous 
for better performance.

\bibliographystyle{IEEEtran} 
\bibliography{icdm2016}

\begin{thebibliography}{10}
\providecommand{\url}[1]{#1}
\csname url@samestyle\endcsname
\providecommand{\newblock}{\relax}
\providecommand{\bibinfo}[2]{#2}
\providecommand{\BIBentrySTDinterwordspacing}{\spaceskip=0pt\relax}
\providecommand{\BIBentryALTinterwordstretchfactor}{4}
\providecommand{\BIBentryALTinterwordspacing}{\spaceskip=\fontdimen2\font plus
\BIBentryALTinterwordstretchfactor\fontdimen3\font minus
  \fontdimen4\font\relax}
\providecommand{\BIBforeignlanguage}[2]{{%
\expandafter\ifx\csname l@#1\endcsname\relax
\typeout{** WARNING: IEEEtran.bst: No hyphenation pattern has been}%
\typeout{** loaded for the language `#1'. Using the pattern for}%
\typeout{** the default language instead.}%
\else
\language=\csname l@#1\endcsname
\fi
#2}}
\providecommand{\BIBdecl}{\relax}
\BIBdecl

\bibitem{active_cancer}
Y.~Liu, ``Active learning with support vector machine applied to gene
  expression data for cancer classification,'' \emph{Journal of Chemical
  Information and Modeling}, vol.~44, no.~6, pp. 1936--1941, 2004.

\bibitem{active_information}
C.~Zhang and T.~Chen, ``An active learning framework for content-based
  information retrieval,'' \emph{{IEEE} Transactions on Multimedia}, vol.~4,
  no.~2, pp. 260--268, 2002.

\bibitem{Survey}
B.~Settles, ``Active learning literature survey,'' University of
  Wisconsin--Madison, Computer Sciences Technical Report 1648, 2009.

\bibitem{ALBL}
W.~Hsu and H.~Lin, ``Active learning by learning,'' in \emph{Proceedings of the
  Twenty-Ninth {AAAI} Conference on Artificial Intelligence, (AAAI) 2015},
  2015, pp. 2659--2665.

\bibitem{survey_transfer}
S.~J. Pan and Q.~Yang, ``A survey on transfer learning,'' \emph{{IEEE} Trans.
  Knowl. Data Eng.}, vol.~22, no.~10, pp. 1345--1359, 2010.

\bibitem{AcTrak}
W.~Daelemans, B.~Goethals, and K.~Morik, ``Actively transfer domain
  knowledge,'' in \emph{Machine Learning and Knowledge Discovery in Databases,
  European Conference, (ECML/PKDD) 2008}, vol. 5212.\hskip 1em plus 0.5em minus
  0.4em\relax Springer, 2008.

\bibitem{transfer_active_stream}
D.~C. Kale and Y.~Liu, ``Accelerating active learning with transfer learning,''
  in \emph{Proceedings of the 2013 {IEEE} 13th International Conference on Data
  Mining}, 2013, pp. 1085--1090.

\bibitem{HATL}
D.~C. Kale, M.~Ghazvininejad, A.~Ramakrishna, J.~He, and Y.~Liu, ``Hierarchical
  active transfer learning,'' in \emph{Proceedings of the 2015 {SIAM}
  International Conference on Data Mining}, 2015, pp. 514--522.

\bibitem{never_ending_learning}
T.~M. Mitchell, W.~W. Cohen, E.~R.~H. Jr., P.~P. Talukdar, J.~Betteridge,
  A.~Carlson, B.~D. Mishra, M.~Gardner, B.~Kisiel, J.~Krishnamurthy, N.~Lao,
  K.~Mazaitis, T.~Mohamed, N.~Nakashole, E.~A. Platanios, A.~Ritter, M.~Samadi,
  B.~Settles, R.~C. Wang, D.~T. Wijaya, A.~Gupta, X.~Chen, A.~Saparov,
  M.~Greaves, and J.~Welling, ``Never-ending learning,'' in \emph{Proceedings
  of the Twenty-Ninth {AAAI} Conference on Artificial Intelligence, (AAAI)
  2015}, 2015, pp. 2302--2310.

\bibitem{life_long_sentiment}
Z.~Chen, N.~Ma, and B.~Liu, ``Lifelong learning for sentiment classification,''
  in \emph{Proceedings of the 53rd Annual Meeting of the Association for
  Computational Linguistics and the 7th International Joint Conference on
  Natural Language Processing of the Asian Federation of Natural Language
  Processing, (ACL) 2015}, 2015, pp. 750--756.

\bibitem{life_long_system}
P.~Ruvolo and E.~Eaton, ``Ella: An efficient lifelong learning algorithm,'' in
  \emph{Proceedings of the 30th International Conference on Machine Learning,
  (ICML) 2013}, 2013, pp. 507--515.

\bibitem{LinUCB}
W.~Chu, L.~Li, L.~Reyzin, and R.~E. Schapire, ``Contextual bandits with linear
  payoff functions,'' in \emph{Proceedings of the 14th International Conference
  on Artificial Intelligence and Statistics, (AISTATS) 2011}, 2011, pp.
  208--214.

\bibitem{uncertainty_start}
D.~D. Lewis and W.~A. Gale, ``A sequential algorithm for training text
  classifiers,'' in \emph{Proceedings of the 17th Annual International
  {ACM-SIGIR} Conference on Research and Development in Information
  Retrieval.}, 1994, pp. 3--12.

\bibitem{uncertainty_svm}
S.~Tong and D.~Koller, ``Support vector machine active learning with
  applications to text classification,'' \emph{Journal of Machine Learning
  Research}, vol.~2, pp. 45--66, 2001.

\bibitem{hint_svm}
C.-L. Li, C.-S. Ferng, and H.-T. Lin, ``Active learning using hint
  information,'' \emph{Neural Computation}, vol.~27, no.~8, pp. 1738--1765,
  August 2015.

\bibitem{representative}
Z.~Xu, K.~Yu, V.~Tresp, X.~Xu, and J.~Wang, ``Representative sampling for text
  classification using support vector machines,'' in \emph{Proceedings of the
  25th European Conference on Information Retrieval Research}, vol. 2633, 2003,
  pp. 393--406.

\bibitem{DUAL}
P.~Donmez, J.~G. Carbonell, and P.~N. Bennett, ``Dual strategy active
  learning,'' in \emph{Proceedings of the 18th European Conference on Machine
  Learning, (ECML) 2007}, 2007, pp. 116--127.

\bibitem{QUIRE}
S.~Huang, R.~Jin, and Z.~Zhou, ``Active learning by querying informative and
  representative examples,'' in \emph{Advances in Neural Information Processing
  Systems, (NIPS) 2010}, 2010, pp. 892--900.

\bibitem{COMB}
Y.~Baram, R.~El{-}Yaniv, and K.~Luz, ``Online choice of active learning
  algorithms,'' in \emph{Proceedings of the 20th International Conference on
  Machine Learning, (ICML) 2003}, 2003, pp. 19--26.

\bibitem{bandit_start}
P.~Auer, N.~Cesa{-}Bianchi, Y.~Freund, and R.~E. Schapire, ``The nonstochastic
  multiarmed bandit problem,'' \emph{{Society of Industrial and Applied
  Mathematics} J. Comput.}, vol.~32, no.~1, pp. 48--77, 2002.

\bibitem{EXP4.P}
A.~Beygelzimer, J.~Langford, L.~Li, L.~Reyzin, and R.~E. Schapire, ``Contextual
  bandit algorithms with supervised learning guarantees,'' in \emph{Proceedings
  of the 14th International Conference on Artificial Intelligence and
  Statistics, (AISTATS) 2011}, 2011, pp. 19--26.

\bibitem{LinRel}
P.~Auer, ``Using confidence bounds for exploitation-exploration trade-offs,''
  \emph{Journal of Machine Learning Research}, vol.~3, pp. 397--422, 2002.

\bibitem{thompson}
S.~Agrawal and N.~Goyal, ``Thompson sampling for contextual bandits with linear
  payoffs,'' in \emph{Proceedings of the 30th International Conference on
  Machine Learning, (ICML) 2013}, 2013, pp. 127--135.

\bibitem{LinUCB_news}
L.~Li, W.~Chu, J.~Langford, and R.~E. Schapire, ``A contextual-bandit approach
  to personalized news article recommendation,'' in \emph{Proceedings of the
  19th International Conference on World Wide Web, (WWW) 2010}, 2010, pp.
  661--670.

\bibitem{bias_reg}
W.~Kienzle and K.~Chellapilla, ``Personalized handwriting recognition via
  biased regularization,'' in \emph{Proceedings of 23rd International
  Conference on Machine Learning, (ICML) 2006}, 2006, pp. 457--464.

\bibitem{liblinear}
R.-E. Fan, K.-W. Chang, C.-J. Hsieh, X.-R. Wang, and C.-J. Lin, ``{LIBLINEAR}:
  A library for large linear classification,'' \emph{Journal of Machine
  Learning Research}, vol.~9, pp. 1871--1874, 2008.

\bibitem{UCI}
M.~Lichman, ``{UCI} machine learning repository,'' 2013.

\bibitem{mtl_1}
Z.~Kang, K.~Grauman, and F.~Sha, ``Learning with whom to share in multi-task
  feature learning,'' in \emph{Proceedings of the 28th International Conference
  on Machine Learning, (ICML) 2011}, 2011, pp. 521--528.

\bibitem{mtl_2}
A.~Kumar and H.~D. III, ``Learning task grouping and overlap in multi-task
  learning,'' in \emph{Proceedings of the 29th International Conference on
  Machine Learning, (ICML) 2012}, 2012.

\end{thebibliography}

\end{document}